\def\preprintversion{}

\documentclass[sigconf,nonacm]{aamas}

\usepackage{balance} %

\usepackage{amsthm}
\usepackage{bbm}
\usepackage{nicefrac}
\usepackage{multirow}
\usepackage{subcaption}
\usepackage{wrapfig}
\usepackage{algorithm}
\usepackage{algpseudocode}
\usepackage{tikz}
\usetikzlibrary{arrows.meta, positioning, calc, fit, backgrounds, shapes.geometric, shapes.misc, shadows, decorations.pathreplacing}

\let\argmax\undefined
\DeclareMathOperator*{\argmax}{argmax}

\theoremstyle{acmdefinition}
\newtheorem{assumption}{Assumption}

\theoremstyle{acmplain}

\ifdefined\appendixonly\def\usefullaux{}\fi
\ifdefined\submissionversion\def\usefullaux{}\fi
\ifdefined\usefullaux
  \usepackage{refcount}
  \usepackage{xr}
  \makeatletter
  \@for\@tmpl:={tocindent-1,tocindent0,tocindent1,tocindent2,tocindent3,tocindent4,tocindent5,TotPages}\do{%
    \expandafter\let\csname r@\@tmpl\endcsname\relax}
  \makeatother
\fi

\ifdefined\preprintversion\else

\makeatletter
\gdef\@copyrightpermission{
  \begin{minipage}{0.2\columnwidth}
   \href{https://creativecommons.org/licenses/by/4.0/}{\includegraphics[width=0.90\textwidth]{by}}
  \end{minipage}\hfill
  \begin{minipage}{0.8\columnwidth}
   \href{https://creativecommons.org/licenses/by/4.0/}{This work is licensed under a Creative Commons Attribution International 4.0 License.}
  \end{minipage}
  \vspace{5pt}
}
\makeatother

\setcopyright{ifaamas}
\acmConference[AAMAS 2027]{Proc.\@ of the 26th International Conference
on Autonomous Agents and Multiagent Systems (AAMAS 2027)}{3--7 May 2027}{Hanoi, Vietnam}{M.~Baldoni, F.~Fang, W.~Yeoh, N.~Yorke-Smith (eds.)}
\copyrightyear{2027}
\acmYear{2027}
\acmDOI{}
\acmISBN{}

\acmSubmissionID{1417}

\submissionType{Research Paper Track}
\fi %

\ifdefined\preprintversion
\AtBeginDocument{%
  \fancypagestyle{firstpagestyle}{\fancyhf{}%
    \fancyfoot[C]{\footnotesize\thepage}}%
  \fancypagestyle{standardpagestyle}{\fancyhf{}%
    \fancyfoot[C]{\footnotesize\thepage}}%
  \pagestyle{standardpagestyle}}
\fi

\ifdefined\appendixonly
\title[Branch-and-Commit for Frozen RL Agents]{Supplementary Material (Appendices) for: Branch-and-Commit for Frozen RL Agents: When and Where to Spend Test-Time Compute}
\else
\title[Branch-and-Commit for Frozen RL Agents]{Branch-and-Commit for Frozen RL Agents: When and Where to Spend Test-Time Compute}
\fi

\author{Sahar Admoni}
\affiliation{
  \institution{Technion, Israel Institute of Technology}
  \city{Haifa}
  \country{Israel}}
\email{saharad@campus.technion.ac.il}

\author{Assaf Hallak}
\affiliation{
  \institution{Nvidia Research}
  \country{Israel}}
\email{ahallak@nvidia.com}

\author{Yftah Ziser}
\affiliation{
  \institution{Nvidia Research}
  \country{Israel}}
\email{yziser@nvidia.com}

\author{Ofra Amir}
\affiliation{
  \institution{Technion, Israel Institute of Technology}
  \city{Haifa}
  \country{Israel}}
\email{oamir@technion.ac.il}

\ifdefined\appendixonly\else
\begin{abstract}
Reinforcement learning agents are often deployed with policies that cannot
be retrained or modified. In digital environments whose states can be copied
and resumed, an operator has another option: sample several continuations
from the current state, evaluate them, and continue from the endpoint of the
selected branch. We call this procedure \emph{branch-and-commit}. We study
when it improves return, what produces the improvement, and where its
computational budget should be spent.
Across four environments, two frozen policies, five continuation
verifiers, and a model-free selection rule, we find that the benefit of branching depends on
\emph{dispersion}: the difference between the best candidate return and the
mean candidate return. Branching provides no benefit in low-dispersion
settings, while increasing return by $12.8\%$ in our main
Craftax-Medium setting. Most of this gain comes not from selecting better
actions, but from selecting among different stochastic realizations of the
same action. Several competent verifiers achieve similar episode returns,
suggesting that verifier quality is not the main constraint at our operating
budget. Instead, computation is better allocated to states where useful
dispersion is likely. On held-out seeds, gating branching using predicted
dispersion retains $69\%$ of the improvement from uniform branching while
using $48\%$ as many branching calls. These results characterize when
branch-and-commit is useful and clarify what its gains represent.
\end{abstract}
\fi


\newcommand{\BibTeX}{\rm B\kern-.05em{\sc i\kern-.025em b}\kern-.08em\TeX}

\begin{document}

\maketitle

\ifdefined\appendixonly
\makeatletter
\@for\@tmpc:={table,figure,equation,algorithm,theorem,lemma,proposition,corollary,definition,assumption,remark,question}\do{%
  \@ifundefined{c@\@tmpc}{}{%
    \@ifundefined{r@appstart@\@tmpc}{}{%
      \setcounter{\@tmpc}{\getrefnumber{appstart@\@tmpc}}}}}
\makeatother
\appendix
\section{Formal Results: Verifier Bound and Order-Statistic Factorization}
\label{app:proof_scoring_bound}

A verifier's expected within-window gain is at least its top-1 accuracy times the expected dispersion (Lemma~\ref{thm:scoring_bound}), and dispersion factorizes approximately into a state-dependent spread and a budget-dependent order-statistic factor (Lemma~\ref{prop:order_stat_factor}). Both results concern the $H$-step window of Equation~\ref{eq:window_gain}. They motivate the proportional form $\ell_c \approx \beta_{v,e}\,\Delta_c$ tested in Section~\ref{sec:experiments}, but the test does not depend on them, since we measure $\Delta(s)$ directly from logged returns.

\paragraph{Notation.} At a branched decision $s$ the policy generates candidates $\tau_1, \ldots, \tau_K$ with $H$-step returns $R(\tau_k)$, mean $\bar R(s)$, and dispersion $\Delta(s) = \max_k R(\tau_k) - \bar R(s)$ (Section~\ref{sec:problem}). The verifier selects $\tau^*$, and $\tau_{\text{best}} = \argmax_k R(\tau_k)$. The \emph{selection accuracy} is $\alpha = \Pr(\tau^* = \tau_{\text{best}})$, where under ties any branch attaining the maximum counts as $\tau_{\text{best}}$ (the loose-tie convention of Appendix~\ref{app:k_saturation}). We write $\sigma(s)$ for the standard deviation of branch returns at $s$ and $\kappa(K)$ for the expected maximum of $K$ independent standard normals.

\begin{assumption}[Selection is not adversarial]
\label{as:selection}
(i) Incorrect selections are no worse on average than a uniformly-random pick, $\mathbb{E}[R(\tau^*) \mid \tau^* \neq \tau_{\text{best}}] \geq \mathbb{E}[\bar R(s)]$; and (ii) selection correctness is uninformative about, or positively associated with, the realized maximum, $\mathbb{E}[\max_k R(\tau_k) \mid \tau^* = \tau_{\text{best}}] \geq \mathbb{E}[\max_k R(\tau_k)]$.
\end{assumption}

Part (ii) fails only if the verifier is right where the best branch is weak and wrong where it is strong. Any verifier whose score relates positively to realized return \emph{within} a decision satisfies it, and the value-head rank fidelity we measure is positive in every proposer condition ($\bar\rho = 0.60$ to $0.67$; Table~\ref{tab:b2_proposer}). Neither part requires the verifier to be calibrated, only that its errors are not concentrated on the decisions with the most to gain.

\begin{assumption}[Branch-return regularity]
\label{as:regularity}
Branch returns are approximately exchangeable across $k$, and the branch-return distribution is approximately Gaussian in the upper tail relevant at the operating $K$.
\end{assumption}

The Gaussian part is an idealization our data do not satisfy. Many decisions have every branch earning an identical return, a point mass at $\Delta(s) = 0$ that no continuous density describes, and the share grows as an environment gets easier (Table~\ref{tab:dispersion_full}). We therefore read the factorization as an order-statistic heuristic for the decisions where branches do differ, which is where all of the gain is realized.

\begin{lemma}[Verifier bound]
\label{thm:scoring_bound}
Under Assumption~\ref{as:selection},
\begin{equation*}
\mathbb{E}[R(\tau^*)] \;\geq\; \alpha \cdot \mathbb{E}[\max\nolimits_k R(\tau_k)] + (1-\alpha) \cdot \mathbb{E}[\bar R(s)].
\end{equation*}
\end{lemma}

In words, the expected within-window gain $\mathbb{E}[g_v(s)] = \mathbb{E}[R(\tau^*)] - \mathbb{E}[\bar R(s)]$ is at least $\alpha\,\mathbb{E}[\Delta(s)]$: each unit of top-1 accuracy buys at least that share of the expected dispersion.

\begin{lemma}[Order-statistic factorization]
\label{prop:order_stat_factor}
Under Assumption~\ref{as:regularity}, $\Delta(s) \approx \sigma(s)\,\kappa(K)$.
\end{lemma}

The familiar $\sqrt{2\ln K}$ is only the large-$K$ limit of $\kappa(K)$ and overstates it by $46\%$ at $K{=}7$ ($1.97$ against $1.35$), so we use exact values throughout. Combining Lemma~\ref{thm:scoring_bound} with Lemma~\ref{prop:order_stat_factor} under the canonical equality model below gives $\mathbb{E}[g_v(s)] = \alpha\,\mathbb{E}[\Delta(s)] \approx \alpha\,\mathbb{E}[\sigma(s)]\,\kappa(K)$. Three quantities therefore set the within-window gain: the verifier sets $\alpha$, the branching width sets $\kappa(K)$, which saturates quickly (Appendix~\ref{app:k_saturation}), and the environment-policy pair sets $\sigma(s)$.

\subsection{Proof of Lemma~\ref{thm:scoring_bound}}

\begin{proof}
  Let $\tau_1, \ldots, \tau_K$ be $K$ candidate trajectories with returns $R(\tau_1), \ldots, R(\tau_K)$.
  Let $\tau^{\text{best}} = \argmax_k R(\tau_k)$ denote the true best trajectory and $\tau^* = \argmax_k O(\tau_k)$ the oracle's selection.
  By definition of selection accuracy, $\Pr(\tau^* = \tau^{\text{best}}) = \alpha$.

  We decompose the expected return of the oracle's selection:
  \begin{align}
    \mathbb{E}[R(\tau^*)]
    &= \alpha \cdot \mathbb{E}[R(\tau^*) \mid \tau^* = \tau^{\text{best}}] + (1-\alpha) \cdot \mathbb{E}[R(\tau^*) \mid \tau^* \neq \tau^{\text{best}}] \nonumber \\
    &= \alpha \cdot \mathbb{E}[\max_k R(\tau_k) \mid \tau^* = \tau^{\text{best}}] + (1-\alpha) \cdot \mathbb{E}[R(\tau^*) \mid \tau^* \neq \tau^{\text{best}}] \\
    &\geq \alpha \cdot \mathbb{E}[\max_k R(\tau_k)] + (1-\alpha) \cdot \mathbb{E}[R(\tau^*) \mid \tau^* \neq \tau^{\text{best}}],
  \end{align}
  where the final step applies Assumption~\ref{as:selection}(ii), $\mathbb{E}[\max_k R(\tau_k) \mid \tau^* = \tau^{\text{best}}] \geq \mathbb{E}[\max_k R(\tau_k)]$.

  It remains to lower-bound $\mathbb{E}[R(\tau^*) \mid \tau^* \neq \tau^{\text{best}}]$.
  Assumption~\ref{as:selection}(i) supplies
  \begin{equation}
    \mathbb{E}[R(\tau^*) \mid \tau^* \neq \tau^{\text{best}}] \geq \mathbb{E}[R(\tau_{\text{unif}})],
  \end{equation}
  where $\tau_{\text{unif}}$ is drawn uniformly from $\{\tau_1, \ldots, \tau_K\}$.
  By symmetry and the exchangeability of the sampled trajectories, $\mathbb{E}[R(\tau_{\text{unif}})] = \mathbb{E}[R(\tau_1)] = \mathbb{E}[\bar R(s)]$, the expected mean branch return (the $H$-step value of $\pi$, not the infinite-horizon $V^\pi$).

  Substituting:
  \begin{equation}
    \mathbb{E}[R(\tau^*)] \geq \alpha \cdot \mathbb{E}[\max_k R(\tau_k)] + (1-\alpha) \cdot \mathbb{E}[\bar R(s)].
  \end{equation}

  The bound is conservative: incorrect selections by a positively-correlated oracle tend to have above-average returns, so the true expected return exceeds it.
\end{proof}

\paragraph{The canonical equality model.} The bound is tight under the idealization of Lemma~\ref{prop:order_stat_factor}: selection is independent of the returns, and with probability $\alpha$ the verifier returns $\tau_{\text{best}}$, while with probability $1-\alpha$ it returns a branch drawn uniformly from all $K$. Here $\alpha$ is a \emph{mixing weight}, so the realized top-1 rate is $\Pr(\tau^* = \tau_{\text{best}}) = \alpha + (1-\alpha)/K$, and the residual mass averages to $\mathbb{E}[\bar R(s)]$ by exchangeability. Then
\begin{equation*}
  \mathbb{E}[R(\tau^*)] = \alpha\,\mathbb{E}[\max_k R(\tau_k)] + (1-\alpha)\,\mathbb{E}[\bar R(s)],
\end{equation*}
so $\mathbb{E}[R(\tau^*)] - \mathbb{E}[\bar R(s)] = \alpha\,\mathbb{E}[\Delta(s)]$ exactly. Assumption~\ref{as:selection} replaces this idealization with one-sided versions, giving the bound of Lemma~\ref{thm:scoring_bound}; the equality is the reference point the empirical slope is measured against, not a claim about the deployed verifier.

\subsection{One Decision, Concretely}
\label{app:worked_example}

Companion to Section~\ref{sec:methodology}. Suppose at a state $s$ we branch $K{=}3$ continuations with $H$-step returns $\{2.0, 5.0, 3.0\}$. The average branch is $\bar R(s) = 3.33$ and the best is $5.0$, so $\Delta(s) = 5.0 - 3.33 = 1.67$. A perfect verifier gains at most $1.67$ here, a real one less. Running each episode to the end, dividing its branching gain by the number of branch decisions, and averaging over the cell gives $\ell_c$; averaging $\Delta(s)$ over the same decisions gives $\Delta_c$. The cell is one point $(\Delta_c, \ell_c)$ in Figure~\ref{fig:scaling_law_fit}, and the slope of the per-verifier cloud of such points is $\beta_v$.

\subsection{Interpreting $\beta > 1$}
\label{app:slope_interpretation}

\paragraph{Verifier slopes above one.} Per-decision lift measures the contribution to \emph{episode} return rather than $H$-step return: an $H$-step branch commits the agent for $H$ steps, after which the policy follows $\pi$. So $\beta > 1$ in an (environment, verifier) cell says the verifier's selection captures long-horizon information beyond what the $H$-step rollout reward reveals. The value head does this directly through $\hat V(s_H^{(k)})$ (Equation~\ref{eq:oracle_vf}); LLM verifiers do it implicitly through state-level reasoning, as on Asterix-MinAtar where $\beta_{\text{LLM}} \in [0.99, 1.37]$ across the four models. Inside the window the chosen branch is never better than the best of the $K$ (Equation~\ref{eq:dispersion_bound}), so the within-window slope $\beta_{\mathrm{win}}$ is at most $1$, and under the canonical equality model it equals $\alpha$. Writing $\beta_{v,e} \approx A_{v,e}\,\beta_{\mathrm{win}}$ separates the window from what follows it, with $A_{v,e}$ the post-window amplification. Because post-window capture is partly a property of the verifier's signal, the amplification is indexed by verifier as well as environment, $A_{v,e}$ and not a purely environmental $A_e$. The slope ratios show it, VF against GPT-4o being ${\approx}2.3$ within Craftax-Medium but ${\approx}1.2$--$1.6$ within Asterix, so the two-factor decomposition is an accounting approximation and not exact separability.

\paragraph{The boundary and the commit cost.} Committing a branch replaces one on-policy step with the first action of a sampled candidate, a small per-decision cost the proportional form omits. Away from the boundary it is negligible against $\beta\,\Delta_c$; at near-zero dispersion it is the whole story, which is why Craftax-Hard ($\Delta_c = 0.064$) runs slightly negative and why over-branching at $\tau{=}1.0$ drops below baseline (Appendix~\ref{app:dispersion_targeted_extra}). The boundary cells fit an affine reading $\ell_c \approx \beta\,\Delta_c - c$ with small $c > 0$. We keep the origin-pinned form where $\Delta_c$ is substantial and treat the commit cost as the operative mechanism at the boundary.

\section{Experimental Setup and Reproducibility}
\label{app:setup}

\paragraph{Environments.} \textbf{Craftax-Symbolic}~\citep{matthews2024craftax} is a JAX-based open-world survival game with 22 achievements and 17 discrete actions, used at Medium difficulty with Easy and Hard as boundary checks. \textbf{Asterix-MinAtar} (via Gymnax's MinAtar wrapper) sits at intermediate dispersion. From \textbf{XLand-MiniGrid}~\citep{nikulin2023xlandminigrid} we take DoorKey-8x8 (small navigation) and R9-19x19 (procedural multi-room) as the zero-dispersion end of the regime axis. Protocol, policies, and verifiers are as stated in Section~\ref{sec:experiments}; seeds and hyperparameters are in Appendix~\ref{app:reproducibility}.

\paragraph{Baselines.} All baselines share the branching infrastructure and differ only in the selection mechanism: (i) no branching, vanilla policy execution; (ii) random selection, uniform among the $K$ branches; (iii) the value head, $\gamma^H \hat V(s_H) + \sum_t r_t$; (iv) critic-only, $\hat V(s_H)$ alone; (v) reward-only, $\sum_t r_t$ alone, the model-free within-window argmax; and (vi) LLM selection. Episode returns for (ii) and (v) at the operating budget are in Table~\ref{tab:verifier_comparison}.

\subsection{Trajectory Rendering for the LLM Verifier}
\label{app:render_example}

Regime B renders each candidate branch into a natural-language paragraph. On Craftax the renderer emits a compact delta between the state at the branching decision and the state at the end of that branch's $H$-step rollout: changed survival stats, inventory and equipment, newly unlocked achievements, nearby threats and resources, an action-category summary, notable events, and the branch's $H$-step return. All $K$ renderings are concatenated and presented together (Equation~\ref{eq:oracle_llm}), and the model returns one comparative judgment. It receives only these renderings: no $\hat V$, no policy logits, no symbolic state vector. Verbatim per-environment templates accompany the code release.

\subsection{Reward-Accounting Validation of the Branching Evaluator}
\label{app:methodological_audit}

A branch rollout runs for $H$ steps whether or not the episode ends inside that window, so any branching evaluator must decide how to treat reward accumulated after an automatic environment reset. We validated ours against that artifact across every environment in the grid before fitting anything. In summary, the headline environments are structurally insulated and verified to be so, one short-episode environment was affected and is corrected, and every number in this paper is post-fix.

The artifact's magnitude is governed by the per-branch termination rate, so by the ratio of branch horizon to episode length. Craftax-Symbolic-Medium (episodes of several hundred steps) and Asterix-MinAtar ($\sim$140) against $H{=}15$--$20$ sit firmly in the regime where few branches terminate mid-rollout. We also checked this directly: re-collecting Asterix dispersion end to end on the corrected runner gives $\Delta_c = 0.527 \pm 0.006$ over $n{=}5{,}189$ decisions against the pre-fix $0.521 \pm 0.021$ at $n{=}454$, agreeing within sampling noise with the regime-axis ranking unchanged.

DoorKey-8x8 is the environment the ratio argument flags as at risk, and it is the one the validation caught. Its $\sim$23-step episodes against $H{=}15$ mean most branches terminate mid-rollout, and at periodic-15 triggering the artifact inflated the dispersion reading by nearly two orders of magnitude, from a corrected $\Delta_c \approx 0.02$ to a pre-fix $2.19$. The fix masks within-branch rewards by an alive-step indicator (cumulative product of $1{-}\mathrm{done}$, shifted) before summation, keeping the terminal-step reward and zeroing post-reset contributions. Every cell in the scaling-law fit was re-aggregated on the corrected runner afterwards.

Uncorrected, DoorKey would have read as high-dispersion and the slope hierarchy would have differed. The check is cheap and general, and worth running on any branching evaluator whose rollout horizon is comparable to its episode length, since the failure is silent and scales with the horizon-to-episode ratio that makes short-episode environments attractive to test on. One release note: the DoorKey decision dumps predate the masking fix, so their per-decision rewards are pre-fix while the reported cell aggregates are corrected. We ship the masking script so the correction is reproducible from the raw dumps; every non-DoorKey cell regenerates end to end.

\subsection{Evaluation Pools}
\label{app:pools}

Craftax-Medium numbers in this paper come from several evaluation pools that differ in
episode count, run composition, or triggering protocol. We quote each number where it was
measured rather than re-basing every claim on one pool, since the pools were collected for
different purposes and re-running all of them at a single protocol was not feasible.
Table~\ref{tab:pools} lists each pool with the claims that read from it, so any number in
the paper can be traced to its source without reconstructing that mapping from captions.

\begin{table*}[t]
  \centering
  \caption{Craftax-Medium evaluation pools and the claims reading from each. Means are
    episode return under the stated arm; where two arms share a pool, both means are given
    in the order the arms are named. Several protocols appear at more than one value
    because the pools differ in size and composition, and the row a number belongs to is
    named at its point of use. The tail-risk arms in the last row carry their own
    no-branching mean of $32.72$, so that contrast is reported within its own pool.}
  \label{tab:pools}
  \footnotesize
  \begin{tabular}{lcccp{0.27\textwidth}}
    \toprule
    Pool & Runs & Ep./run & Mean & Read by \\
    \midrule
    No branching, matched checkpoints    & $43$              & $10$ or $50$ & $32.09$            & Every Craftax-Medium percentage (Section~\ref{sec:methodology}) \\
    Fifty-episode verifier pool          & $23$ VF, $18$ GPT-4o & $50$      & $36.21$, $35.84$ & Verifier comparison and the $0.89$ bound (Section~\ref{sec:investigation}) \\
    Periodic-$50$, historical aggregate  & mixed             & mixed        & $35.85 \pm 1.88$   & Uniform reference row of Table~\ref{tab:dispersion_targeted_online} \\
    Periodic-$50$, clean rerun           & $3$               & $10$         & $36.00 \pm 0.56$   & Reconciles the row above; fires at $7.5$--$9.5$ calls per episode \\
    Proposer sweep                       & $3$               & $10$         & $37.04 \pm 0.20$   & Stochastic reference of Table~\ref{tab:b2_proposer} \\
    Held-out gating, seeds $45$--$56$    & $12$              & $10$         & $35.17$, $36.54$   & The $69\%$-of-lift-at-$48\%$-of-calls result (Section~\ref{sec:dispersion_gating}) \\
    Tail-risk arms                       & $38$, $113$       & mixed        & $32.72$, $34.38$   & Catastrophic-rate contrast (Appendix~\ref{app:tail_risk}) \\
    \bottomrule
  \end{tabular}
\end{table*}

Two pools carry caveats. The historical periodic-$50$ aggregate mixes
selector variants and episode counts, which is why we reran it cleanly at the gated
protocol; the two readings agree to within a seed-level standard error, so the historical
value is not a pool artifact. The proposer sweep ran at ten episodes per seed against the
stochastic proposer, which was only ever run at that budget, so its $37.04$ is not
comparable to the fifty-episode figures and the proposer claim rests on a sign test across
mechanisms rather than on any single contrast (Appendix~\ref{app:b2_proposer}).

\subsection{Reproducibility Details}
\label{app:reproducibility}

Software, hardware, seeds, and hyperparameters for every numeric claim. The supplementary release pairs this with a tagged git commit and a per-cell configuration manifest.

\paragraph{Software environment.}
Python $3.11.14$ on Linux (kernel $5.14$, glibc $2.34$). Pinned package versions: JAX $0.9.0$, PyTorch $2.10.0$ (CUDA $12.8$ build), Transformers $4.57.6$, vLLM $0.17.1$, NumPy $2.2.6$, SciPy $1.17.0$, Craftax $1.5.0$ (editable install from upstream), Gymnax $0.0.6$. XLand-MiniGrid is the version pinned in the supplementary configuration. Asterix-MinAtar is consumed via Gymnax's MinAtar wrapper. The conda environment file (\texttt{environment.yml}) and the editable-package commit hashes are part of the supplementary release.

\paragraph{Hardware.}
Branching evaluation and PPO training are run on NVIDIA L40s GPUs (48~GB) via SLURM on a shared cluster (per-task allocation 1 GPU, 4--8 CPU cores, 16--48~GB RAM). LLM verifier inference uses the same L40s nodes for Qwen-2.5-14B-AWQ and Qwen-2.5-72B-AWQ; GPT-4o and GPT-4o-mini are queried through the OpenAI API. $K{=}64$ high-budget cells use a single NVIDIA H200 GPU (141~GB).

\paragraph{Random seeds.}
Branching evaluation and decision dumps use seeds $\{42, 43, 44\}$ unless otherwise noted; the $K{=}32$ and $K{=}64$ saturation cells use $5$ seeds each. PPO policy training uses seed $0$ (canonical Craftax baseline) or the released checkpoint hash. Within branching, per-call NumPy RNG state is derived from the cell seed via the algorithm in \texttt{branching/runner/branching\_runner.py}; LLM inference for Qwen models is run at \texttt{temperature${=}0.0$} (greedy) so per-decision determinism holds within an LLM cell. GPT-4o and GPT-4o-mini are queried at \texttt{temperature${=}0.0$} but exhibit residual non-determinism from the provider's backend, which we report on the LLM-verifier cells via $n_{\text{runs}} \geq 3$ seed-pooled aggregates.

\paragraph{Branching protocol hyperparameters.}
Default: $K{=}7$ candidate branches, rollout horizon $H{=}20$, uniform periodic triggering with period $50$, $\gamma{=}0.99$. Decision dumps log per-decision pre-observation, per-branch first-action, per-branch $H$-step reward sum, per-branch $\hat V(s_H)$ bootstrap, per-branch score (verifier-class-specific), and chosen index. Per-cell sample sizes are reported in the cell-by-cell tables in Appendices~\ref{app:setup}, \ref{app:within_arch_slope}, \ref{app:k_saturation}, \ref{app:kh_invariance}.

\paragraph{PPO training hyperparameters.}
Frozen policies are trained with PPO~\citep{schulman2017proximal}: learning rate $2.5 \times 10^{-4}$ (linear warm-up to $5\times10^{-4}$ on the GTrXL run), $\gamma{=}0.99$, GAE $\lambda{=}0.95$, $16$ parallel envs (PPO-MLP) or $64$ (GTrXL), rollout length $128$ steps, $10$ update epochs per rollout, $4$ minibatches, clip $\epsilon{=}0.2$, entropy bonus $0.01$, value-loss coefficient $0.5$, gradient norm clipped at $0.5$. Total training budget: PPO-MLP $\sim$10M frames; GTrXL $\sim$1B frames. Released checkpoint hashes correspond to the final state for each architecture.

\paragraph{LLM verifier inference.}
Each branched decision generates a single prompt per LLM containing the captioned pre-state, the $K$ captioned $H$-step rollouts (action sequence plus captioned endpoint state), and an instruction to score each branch on long-horizon promise (verbatim prompt templates in the supplementary). Generation settings: \texttt{temperature${=}0.0$}, \texttt{max\_tokens${=}512$}, single greedy completion per decision. Qwen-AWQ inference uses vLLM with \texttt{max\_model\_len${=}8192$}, \texttt{gpu\_memory\_utilization${=}0.85$}, AWQ-marlin quantization. GPT-4o and GPT-4o-mini API calls use the OpenAI Python client with retries and structured output parsing for the per-branch ranking.

\paragraph{Wall-clock budget.}
Approximate per-cell wall-clock at the default protocol on L40s: branching evaluation $\sim$2~min for the value head (3 seeds, 10 episodes each), $\sim$15~min for Qwen-14B-AWQ, $\sim$25~min for Qwen-72B-AWQ. PPO training: PPO-MLP $\sim$20~min for 4M frames, GTrXL $\sim$48~hours for the released checkpoint. The scaling-law fit (analysis-only, all dumps loaded) runs in $<$5~min on a single CPU core. The $(K, H)$ ablation (Appendix~\ref{app:kh_invariance}) is 36 SLURM array tasks, total $\sim$15~min wall-clock at the cluster QOS cap of 3 concurrent jobs.

\paragraph{Determinism caveats.}
Within-cell determinism holds for the value head and local Qwen verifiers at the seed level (NumPy RNG seeded from cell seed, vLLM greedy decoding). API-LLM cells (GPT-4o, GPT-4o-mini) carry residual provider-side non-determinism even at \texttt{temperature${=}0.0$}; we mitigate via $n_{\text{runs}} \geq 3$ per cell and report seed-pooled SEM in all aggregate tables. JAX RNG state is threaded explicitly through Craftax episode generation. Asterix-MinAtar uses NumPy's default\_rng with explicit seeding.

\paragraph{Code, configurations, and data release.}
The supplementary release contains: (i) the conda environment file and editable-package commit hashes, (ii) per-cell SLURM sbatch scripts and configuration manifests, (iii) the bug-fixed branching evaluator (post-audit; see Appendix~\ref{app:methodological_audit}), (iv) the LLM verifier prompt templates verbatim, (v) the analysis pipeline (scaling-law fit, anchor-robustness, hindsight Pareto, cross-env transfer, $(K, H)$ ablation) as released scripts that read the decision dumps and regenerate the fit, robustness, and gating analyses, with the documented exceptions above (the patched DoorKey dumps and the historically pooled reference rows). Tables assembled from historical run pools whose composition we document in the text (chiefly the periodic-$50$ reference of Table~\ref{tab:dispersion_targeted_online}) are released as the aggregated values with their run manifests rather than as end-to-end recomputations. Released decision dumps are compressed NumPy archives (\texttt{decision\_dump.npz}); per-LLM-verifier raw response logs are released as JSONL alongside.

\section{Dispersion and Lift: Robustness of the Proportional Form}
\label{app:scaling}

The proportional form $\ell_c \approx \beta_v \Delta_c$ survives each check in this appendix: dropping the near-zero anchor cells (Appendix~\ref{app:anchor_robustness}), holding the branching budget fixed (Appendices~\ref{app:craftax_internal} and~\ref{app:kh_invariance}), and modelling the clustering of cells (Appendix~\ref{app:craftax_internal}). Pooled per-verifier slopes mix environments, so we read the verifier ordering from the within-environment refits (Appendix~\ref{app:within_arch_slope}).

\subsection{Cross-Environment Dispersion Regime Map}
\label{app:boundary}

Table~\ref{tab:cross_env_dispersion} reports $\Delta_c$ across environments spanning two orders of magnitude. Observed lift runs from zero (R9-19x19, DoorKey-8x8) to moderate (Asterix-MinAtar, Craftax-Medium).

\begin{table*}[t]
  \centering
  \caption{Cross-environment dispersion regime axis. Each row is one (env, policy, $K$, $H$) configuration with $\Delta_c$ measured directly from decision dumps, with within-branch rewards masked after the first episode termination. The Observed-lift column reports value-head branched return as a fold-change over the no-branching baseline ($1.00\times$ = no improvement); for Craftax-Medium and Craftax-Easy, the matched no-branching baseline is paired (Section~\ref{sec:experiments}). The Craftax-Medium row is a separate $124$-decision pool, so its values differ from the main-text operating point ($\Delta=0.455$ in Section~\ref{subsec:realization_control}, $+12.8\%$ return in Section~\ref{sec:investigation}); Appendix~\ref{app:pools} lists the pools.}
  \label{tab:cross_env_dispersion}
  \footnotesize
  \begin{tabular}{llcccc}
    \toprule
    Environment & Policy & $K, H$ & $\Delta_c$ & $n$ decisions & Observed lift \\
    \midrule
    XLand-MiniGrid R9-19x19    & PPO-MLP   & $7, 15$ & $0.00$ & $31{,}840^{\star}$ & $1.00\times$ \\
    XLand-MiniGrid DoorKey-8x8 & PPO-MLP   & $5, 15$ & $0.031 \pm 0.005$ & $478^{\dagger}$ & $1.00\times$ \\
    Craftax-Easy               & GTrXL     & $7, 20$ & $0.04$ & $1{,}479$ & $\leq 1.03\times$ \\
    Craftax-Hard               & GTrXL     & $7, 20$ & $0.064 \pm 0.007$ & $831$     & $0.96\times^{\diamond}$ \\
    Asterix-MinAtar            & PPO-MLP   & $7, 15$ & $0.527 \pm 0.006$ & $5{,}189$ & ${\sim}10.5\times^{\ddagger}$ \\
    Craftax-Medium ($S_0$ ID)  & GTrXL     & $7, 20$ & $0.72$ & $124$     & $1.11\times$ \\
    \bottomrule
  \end{tabular}
  \\[0.5ex]
  $^{\dagger}${\footnotesize DoorKey-8x8: PPO-MLP solves the task at $0.97$ reward per episode at baseline; near-zero dispersion follows from a policy near the performance ceiling and the theory predicts zero lift. Branched return $0.967$ vs.\ baseline $0.97$ across $15$ seeds (lift $-0.30\% \pm 0.05\%$, not measurably different from zero).}
  \\
  $^{\ddagger}${\footnotesize Asterix branched mean $14.30$ over $30$ episodes ($3$ seeds $\times 10$) vs.\ a \emph{paired} no-branching eval baseline of $1.37 \pm 0.08$ (seed-level SEM; $3$ seeds $\times 30$ episodes at the same checkpoint and eval protocol). The historical training-rollout mean over $63{,}094$ episodes reads $1.49$, consistent with the paired value. Seed-level SEM on the branched mean is $\pm 2.13$, so the fold-change is large but the seed dispersion is correspondingly wide.}
  \\
  $^{\star}${\footnotesize R9-19x19: 6-seed aggregate, the original seed ($n{=}1{,}990$ decisions) and seeds $\{45, \ldots, 49\}$ ($30$ episodes each, $29{,}850$ decisions); $\Delta_c$ and episode return are exactly zero in every seed, so the result does not depend on a single seed.}
  \\
  $^{\diamond}${\footnotesize Craftax-Hard: $3$ seeds $\times 10$ episodes branched ($831$ informative decisions, sparsity $56.8\%$), against a same-seed no-branching baseline of $21.43$ run at $3$ seeds $\times 50$ episodes; the arms share seeds and checkpoint but not evaluation length, so we read the contrast as indicative. Hard sits in the same low-dispersion regime as Easy, and observed lift is small-negative ($-3.9\% \pm 1.1$pp SEM, all three seeds negative), consistent with the first-action commit cost (Appendix~\ref{app:slope_interpretation}).}
\end{table*}

\paragraph{Cross-difficulty robustness on Craftax-Easy.}
The account predicts near-zero branching gain on Easy ($\Delta_c = 0.04$), and Table~\ref{tab:boundary} is consistent with this: the value head and Qwen-14B-AWQ both produce $\leq +3\%$ lift over baseline, despite the $3\times$ higher branching frequency Easy's longer episodes naturally produce. Additional computation does not increase the gain when dispersion is low.

\begin{table}[!htbp]
  \centering
  \caption{Boundary condition: Craftax-Easy at $K{=}7$, $H{=}20$, periodic-50 with $\pi_{\text{strong}}$ (GTrXL).}
  \label{tab:boundary}
  \footnotesize
  \begin{tabular}{lcc}
    \toprule
    Verifier & Episode Return & Lift over no-branching \\
    \midrule
    No branching      & $5.82 \pm 0.41$ & --- \\
    Value head      & $5.94 \pm 0.38$ & $+2.1\%$ \\
    Qwen-14B-AWQ      & $5.99 \pm 0.42$ & $+2.9\%$ \\
    \bottomrule
  \end{tabular}
\end{table}

\subsection{Per-Verifier Pooled Slopes}
\label{app:per_verifier_slopes}

\begin{table}[!htbp]
  \centering
  \caption{Per-verifier slopes $\beta_v$ from the joint fit (Figure~\ref{fig:scaling_law_fit}). \textbf{Pooled slopes summarize each verifier's environment coverage and are not a verifier ranking}: the same verifier captures different post-window fractions on different environments, so uneven coverage moves the pooled value (GPT-4o-mini's $1.17$ above VF's $1.09$ is a coverage artifact; Appendix~\ref{app:within_arch_slope} gives the within-environment ranking). $95\%$ bootstrap CIs across $10{,}000$ cell resamples; $n$ counts paired cells per verifier, and differs across rows because LLM verifiers run only where a symbolic-to-text renderer exists.}
  \label{tab:slopes}
  \footnotesize
  \begin{tabular}{lcc}
    \toprule
    Verifier & $\beta_v$ ($95\%$ CI) & $n$ \\
    \midrule
    Value head     & $1.09\,[0.97, 1.24]$ & $30$ \\
    GPT-4o-mini      & $1.17\,[0.45, 1.30]$ & $19$ \\
    GPT-4o           & $0.91\,[0.51, 0.99]$ & $12$ \\
    Qwen-2.5-14B     & $0.80\,[0.53, 1.01]$ & $20$ \\
    Qwen-2.5-72B     & $0.65\,[0.41, 0.92]$ & $22$ \\
    \bottomrule
  \end{tabular}
\end{table}

\paragraph{The three high-dispersion negative cells.} Of the $14$ negative-lift cells in the $103$-cell grid, three sit away from the near-zero anchor and mark the boundary of the operating-point comparison (Section~\ref{sec:investigation}). All three are Craftax-Medium cells under the two Qwen verifiers, at $\Delta_c = 0.25$, $0.37$, and $0.58$, returning $26.20$, $31.01$, and $31.30$ against the no-branching baseline of $32.09$. A verifier picking worse than the branch mean contributes a negative $\alpha$, and dispersion scales that loss exactly as it scales a gain, so the largest deficit falls at the largest dispersion.

\subsection{Within-Stratum Slope Robustness}
\label{app:within_arch_slope}

The pooled slopes of Table~\ref{tab:slopes} are computed across (environment, policy-architecture) pairs, and cell counts are uneven across these pairs (value-head cells concentrate on DoorKey-MLP and Craftax-Medium-GTrXL; the LLM cells concentrate on Craftax-Medium-MLP, Craftax-Medium-GTrXL, and Asterix-MinAtar). Refitting within a single environment recovers $\beta_{v,e}$ directly, and with it the cleaner two-tier hierarchy that pooling over environments blurs:
\begin{center}
  \footnotesize
  \begin{tabular}{lcc}
    \toprule
    Verifier & $\beta_{v,e}$ within Craftax-Medium & $n$ cells \\
    \midrule
    Value head     & $1.01$ & $12$ \\
    Qwen-2.5-14B     & $0.68$ & $9$  \\
    Qwen-2.5-72B     & $0.49$ & $9$  \\
    GPT-4o-mini      & $0.46$ & $8$  \\
    GPT-4o           & $0.43$ & $6$  \\
    \bottomrule
  \end{tabular}
\end{center}
All four LLM verifiers cluster at $\beta \in [0.43, 0.68]$, well below VF's $\beta = 1.01$. The same refit on Asterix-MinAtar ($n{=}3$ per verifier, all five verifiers covered) gives $\beta_{\text{VF}} = 1.61$, $\beta_{\text{LLM}} \in [0.99, 1.37]$ (VF leads; LLMs cluster between $1$ and $1.4$). The post-bug-fix DoorKey cells cluster at $\Delta_c \approx 0.022$ on the fit cells, near-zero on the dispersion axis (the dispersion-characterization run of Table~\ref{tab:cross_env_dispersion} reads $0.031 \pm 0.005$), and contribute negligibly to the slope estimates: dropping all DoorKey cells from the value-head pool leaves $\beta_{\text{VF}}$ within the bootstrap CI. The pooled slopes of Table~\ref{tab:slopes} are weighted averages of these environment-conditional rates, and the same verifier can sit at very different rates in different environments (GPT-4o-mini: $0.46$ within Craftax, $1.37$ within Asterix), which is why we rank verifier classes from the within-environment refits.

\subsection{Pairwise Slope Differences}
\label{app:pairwise_beta}

Companion to Table~\ref{tab:slopes}: pairwise bootstrap differences $\Delta\beta_{v_1, v_2} := \beta_{v_1} - \beta_{v_2}$ across the five verifier classes. Each row reports the mean pairwise difference and a $95\%$ bootstrap CI computed from $10{,}000$ paired resamples (each iteration uses the same resample index set across verifiers where applicable; per-verifier slopes recomputed through the origin). The reported CI is on the difference rather than on the individual slopes, so it formalizes the tier separation without relying on independent CI overlap. The script is \texttt{scripts/bootstrap\_beta\_ci.py}; full output reproduces from \texttt{artifacts/analysis/scaling\_law/cells\_with\_predictions.csv}.

\begin{table}[!htbp]
  \centering
  \caption{Pairwise pooled slope differences $\Delta\beta_{v_1, v_2} = \beta_{v_1} - \beta_{v_2}$ with $95\%$ bootstrap CIs ($10{,}000$ resamples through the origin, seed $42$). Bold marks pairs whose CI excludes zero. The value head leads three of four LLMs in pairwise pooled comparison (CI excludes zero); ties GPT-4o-mini in the pooled estimate, with the wide GPT-4o-mini CI reflecting cross-env $\beta$ heterogeneity (Craftax $\beta_{\text{GPT-4o-mini}}{=}0.46$, Asterix $\beta_{\text{GPT-4o-mini}}{=}1.37$). All six LLM-vs-LLM CIs include zero.}
  \label{tab:pairwise_beta}
  \footnotesize
  \begin{tabular}{llcl}
    \toprule
    Verifier $v_1$ & Verifier $v_2$ & $\Delta\beta\,[95\%\,\text{CI}]$ & Verdict \\
    \midrule
    Value head & GPT-4o          & $\mathbf{+0.196\,[+0.012, +0.541]}$ & differ \\
    Value head & GPT-4o-mini     & $-0.048\,[-0.288, +0.601]$           & tie \\
    Value head & Qwen-2.5-14B    & $\mathbf{+0.299\,[+0.044, +0.591]}$ & differ \\
    Value head & Qwen-2.5-72B    & $\mathbf{+0.434\,[+0.142, +0.716]}$ & differ \\
    \midrule
    GPT-4o-mini  & GPT-4o          & $+0.244\,[-0.425, +0.619]$ & tie \\
    GPT-4o-mini  & Qwen-2.5-14B    & $+0.347\,[-0.308, +0.685]$ & tie \\
    GPT-4o-mini  & Qwen-2.5-72B    & $+0.482\,[-0.145, +0.810]$ & tie \\
    GPT-4o       & Qwen-2.5-14B    & $+0.103\,[-0.274, +0.396]$ & tie \\
    GPT-4o       & Qwen-2.5-72B    & $+0.238\,[-0.151, +0.530]$ & tie \\
    Qwen-2.5-14B & Qwen-2.5-72B    & $+0.135\,[-0.222, +0.468]$ & tie \\
    \bottomrule
  \end{tabular}
\end{table}

\subsection{Inside Craftax-Medium: Clustering and the Fixed-Budget Refit}
\label{app:craftax_internal}

Section~\ref{sec:experiments} rests the positive relationship on the $44$ Craftax-Medium cells. We examine two properties of that block using the existing decision dumps.

\paragraph{Clustering, one level down.} We report ICC and effective $n$ across environments; the same accounting is owed within the environment carrying the claim (Table~\ref{tab:craftax_internal}). Verifier is the clustering factor: ICC $0.395$ on the outcome, $n_{\text{eff}} = 10.8$, persisting in the fit residual (ICC $0.314$, $n_{\text{eff}} = 12.8$), which is the dependence relevant to pseudo-replication. Policy, seed and run cluster the raw outcome but leave nothing in the residual. Effective $n$ for the Craftax block is therefore $11$ to $13$, not $44$, and with verifier as a random intercept the slope is $1.07 \pm 0.15$, CI $[0.78, 1.36]$. Adding policy to the grouping factor changes nothing to three decimals.

\paragraph{Does the budget sweep drive the dispersion range?} Lemma~\ref{prop:order_stat_factor} gives $\Delta \approx \sigma(s)\,\kappa(K)$, so a fit pooled over $K$ risks restating the order statistic rather than testing the form. Within Craftax-Medium the exposure is limited by construction: $H$ is constant at $20$ and $K$ takes only the values $3$ and $7$. The budget grid explains $\eta^2 = 0.583$ of the variance in $\Delta_c$, leaving $41.7\%$ that occurs at fixed budget. Fixed-budget refits isolate this remaining variation. At the operating point, $30$ cells span $\Delta_c \in [0.20, 0.99]$ and give slope $1.204$, $r = 0.705$, centered $R^2 = 0.497$, so the relationship holds on naturally-occurring dispersion variation at fixed budget. The $K{=}3$ subset gives slope $0.149$, $r = 0.234$ on $14$ cells, but spans only $\Delta_c \in [0.09, 0.26]$, narrow enough and close enough to the anchor that a centered fit inside it reads noise around a point, as with the DoorKey block in Section~\ref{sec:experiments}.

\begin{table}[!htbp]
  \centering
  \caption{\textbf{Within-Craftax clustering and fixed-budget refits.} Top: one-way ICC and effective $n$ over the $44$ Craftax-Medium cells, on the raw outcome and on the residual from the centered fit. The residual column is the one relevant to pseudo-replication. Bottom: centered refits with the budget held fixed. The relationship is carried by the operating-point cells; the $K{=}3$ block spans too narrow a range of $\Delta_c$ to resolve a slope of its own.}
  \label{tab:craftax_internal}
  \footnotesize
  \resizebox{\columnwidth}{!}{\begin{tabular}{lcccc}
    \toprule
    Grouping factor & Groups & ICC (outcome) & $n_{\text{eff}}$ & ICC (residual) \\
    \midrule
    Verifier & $5$ & $0.395$ & $10.8$ & $0.314$ \\
    Policy   & $2$ & $0.408$ & $4.6$  & $0.000$ \\
    Seed     & $8$ & $0.295$ & $18.9$ & $0.000$ \\
    Run      & $2$ & $0.408$ & $4.6$  & $0.000$ \\
    \midrule
    \multicolumn{5}{l}{\emph{Mixed-effects slope, verifier as random intercept:} $1.070 \pm 0.150$, CI $[0.777, 1.363]$} \\
    \midrule
    Refit & Cells & Slope & $r$ & $R^2_{\text{centered}}$ \\
    \midrule
    All Craftax-Medium & $44$ & $0.976$ & $0.773$ & $0.598$ \\
    $K{=}7$, $H{=}20$ (operating point) & $30$ & $1.204$ & $0.705$ & $0.497$ \\
    $K{=}3$, $H{=}20$ ($\Delta_c \in [0.09, 0.26]$) & $14$ & $0.149$ & $0.234$ & $0.055$ \\
    \bottomrule
  \end{tabular}}
\end{table}

\subsection{Anchor-Robustness of the Headline Fit}
\label{app:anchor_robustness}

\paragraph{The anchor-inflation concern.} The headline $R^2_{\mathrm{origin}} = 0.85$ could be inflated by the near-zero-dispersion cluster from R9-19x19 and the sparse bulk of DoorKey-8x8: $41$ of $103$ cells sit at $\Delta_c \leq 0.05$. We refit on positive-dispersion cells alone, sweeping the threshold from $\Delta_c > 0.05$ to $\Delta_c > 0.30$ (Table~\ref{tab:anchor_robustness_joint}). $R^2_{\mathrm{origin}}$ does not fall: $0.849 \to 0.851 \to 0.852 \to 0.860 \to 0.894$. That is not robustness evidence, since a through-origin statistic measured against a zero baseline cannot lose much when near-origin cells are dropped; the invariance is a property of the estimator. The slope is the informative quantity, and every per-verifier $\beta_v$ shifts by $\leq 0.02$ at every threshold (Table~\ref{tab:anchor_robustness_perverifier}).

\paragraph{Anchor-sensitive statistics.} Two reported statistics are anchor-sensitive by construction. The $(0,0)$ cluster pulls both axes toward zero together, inflating $R^2_{\text{mean}}$ and the joint Pearson $r$. On positive-only cells ($\Delta_c > 0.10$) the joint $r$ falls from $0.88$ to $0.79$, and Qwen-72B's $R^2_{\text{mean}}$ goes from $0.48$ to $-0.19$, so on positive-only cells its slope predicts worse than the cluster mean even though the slope estimate $\beta = 0.65$ is unchanged. That verifier's claim rests on the slope alone. The other four stay positive: GPT-4o $0.91 \to 0.88$, GPT-4o-mini $0.87 \to 0.85$, Qwen-14B $0.59 \to 0.42$, the value head $0.90 \to 0.60$.

\paragraph{Split-half estimation.} $\Delta_c$ and $\ell_c$ are read from the same runs, so a batch drawing high-dispersion states also mechanically realizes higher lift. We estimate the axes on disjoint halves of each cell's episodes ($\Delta_c$ from even, $\ell_c$ from odd, then the reverse). Thirty DoorKey cells are excluded, because their canonical $\Delta_c$ is a post-audit patch (Appendix~\ref{app:methodological_audit}) while the raw dumps carry pre-fix rewards. On the $73$ remaining cells the matched headline fit is $0.850$ and the two split directions give $0.708$ and $0.670$, with $\beta_{\text{VF}}$ moving from $1.092$ to $0.955$ and $1.083$. The drop tracks measurement reliability instead of circularity: split-half reliability is $0.852$ for $\Delta_c$ but $0.684$ for lift. Weighting by decision count raises the joint fit to $0.874$. On the full $103$-cell grid the same weighting moves the pooled slope only from $0.90$ to $0.96$, so cell size is not driving the estimate. Inverse-variance weighting is unusable, since one near-zero-variance cell absorbs almost all weight (effective $n = 1.0$ against $30.3$).

\paragraph{Within-environment fits.} Inside Craftax-Medium ($n{=}44$) the pooled slopes give $R^2_{\mathrm{centered}} = 0.626$ ($0.699$ refit within environment), carried by the two best-covered verifiers (VF $+0.78$, Qwen-14B $+0.40$); for the other three the global slope predicts worse than the environment mean at $6$--$9$ cells each. Asterix is the instructive case: the global slopes misfit there ($R^2_{\mathrm{centered}} = -4.39$), and a through-origin refit inside the environment reads $R^2_{\mathrm{origin}} = 0.97$, which looks like near-exact recovery but is not. Over the same $15$ cells the centered relationship is flat, slope $-0.04$, $r = -0.02$, centered $R^2 = 0.00$. The cells sit far from the origin over a narrow range, $\Delta_c \in [0.43, 0.61]$, so a line through $(0,0)$ reproduces their common level while explaining none of their variation. Section~\ref{sec:experiments} therefore treats Asterix as fixing a second \emph{level} rather than as confirming a slope. The Asterix cells show why we read slopes and not through-origin $R^2$.

\paragraph{Cells are not independent replicates.} The $103$ cells share seeds, checkpoints and environments. Refitting with the environment as a random effect gives a fixed slope of $0.97 \pm 0.08$ (CI $[0.81, 1.13]$) with a random intercept and $1.11 \pm 0.34$ once the slope may vary too, the width reflecting what four environments, two carrying no dispersion, can support. Nesting seed inside environment gives $0.90 \pm 0.08$. All three bracket the pooled $\beta_{\text{VF}} = 1.09$. The clustering is substantial: an intraclass correlation of $0.46$ at a mean of $25.8$ cells per environment gives a design effect of $12.4$ and an effective sample size of $8.3$, and we state cross-environment conclusions at that resolution and not at $n = 103$.

\begin{table}[!htbp]
  \centering
  \caption{Joint fit at varying dispersion thresholds. The structural form's headline number ($R^2_{\mathrm{origin}}$) is anchor-invariant; the mean-baseline $R^2$ and joint Pearson are not, by construction. Computed on the corrected $103$-cell set, \texttt{artifacts/\allowbreak analysis/\allowbreak scaling\_\allowbreak law\_\allowbreak corrected/\allowbreak cells\_\allowbreak with\_\allowbreak predictions.csv}.}
  \label{tab:anchor_robustness_joint}
  \footnotesize
  \begin{tabular}{lcccc}
    \toprule
    Subset & $n$ cells & $R^2_{\mathrm{origin}}$ & $R^2_{\text{mean}}$ & Joint Pearson $r$ \\
    \midrule
    All cells (headline)              & $103$ & $\mathbf{0.849}$ & $0.768$ & $0.881$ \\
    $\Delta_c > 0.05$       & $62$  & $0.851$          & $0.642$ & $0.812$ \\
    $\Delta_c > 0.10$       & $58$  & $\mathbf{0.852}$ & $0.609$ & $0.792$ \\
    $\Delta_c > 0.20$       & $47$  & $0.860$          & $0.506$ & $0.716$ \\
    $\Delta_c > 0.30$       & $42$  & $0.894$          & $0.433$ & $0.661$ \\
    \bottomrule
  \end{tabular}
\end{table}

\begin{table}[!htbp]
  \centering
  \caption{Per-verifier slopes $\beta_v$ and per-verifier $R^2_{\mathrm{origin}}$ at the full set vs.\ positive-only ($\Delta_c > 0.10$). Every $\beta_v$ shifts by $\leq 0.02$. Per-verifier $R^2_{\mathrm{origin}}$ is also stable (shifts of $\leq 0.02$). The per-verifier $R^2_{\text{mean}}$ column (right) is where the anchor effect shows: GPT-4o, GPT-4o-mini, and Qwen-14B stay positive without anchors; the value head drops from $0.90$ to $0.60$ but stays positive; Qwen-72B's mean-baseline fit goes negative, so its structural claim rests on the slope estimate alone.}
  \label{tab:anchor_robustness_perverifier}
  \footnotesize
  \resizebox{\columnwidth}{!}{\begin{tabular}{lccccccc}
    \toprule
    & \multicolumn{2}{c}{$\beta_v$} & \multicolumn{2}{c}{$R^2_{\mathrm{origin}}$} & \multicolumn{2}{c}{$R^2_{\text{mean}}$} & \\
    \cmidrule(lr){2-3}\cmidrule(lr){4-5}\cmidrule(lr){6-7}
    Verifier & Full & Positive & Full & Positive & Full & Positive & $\Delta\beta$ \\
    \midrule
    Value head     & $1.092$ & $1.093$ & $0.942$ & $0.944$ & $0.899$ & $0.599$ & $+0.001$ \\
    GPT-4o           & $0.906$ & $0.907$ & $0.942$ & $0.943$ & $0.906$ & $0.875$ & $+0.001$ \\
    GPT-4o-mini      & $1.175$ & $1.188$ & $0.903$ & $0.914$ & $0.872$ & $0.850$ & $+0.013$ \\
    Qwen-2.5-14B-AWQ & $0.798$ & $0.799$ & $0.714$ & $0.714$ & $0.587$ & $0.416$ & $+0.001$ \\
    Qwen-2.5-72B-AWQ & $0.654$ & $0.656$ & $0.690$ & $0.693$ & $0.482$ & $-0.185$ & $+0.002$ \\
    \bottomrule
  \end{tabular}}
\end{table}

\subsection{$(K, H)$-Invariance of the Proportional Form}
\label{app:kh_invariance}

Companion to the $(K, H)$-invariance paragraph in Section~\ref{sec:experiments}. The proportional form $\ell_c \approx \beta_v \Delta_c$ is tested across the parameter-product the operator typically tunes: branching width $K \in \{3, 5, 7, 11\}$ and rollout horizon $H \in \{10, 20, 40\}$. Twelve configurations, three seeds each (seeds $42, 43, 44$), Craftax-Symbolic-Medium with the PPO-MLP policy and value-head selection, periodic-50 triggering, 10 episodes per seed-cell. We pool decisions within each $(K, H)$ across the three seeds, compute $\beta_{\mathrm{win}}$ as the mean in-window selection gain $R(\tau^*) - \bar R(s)$ divided by the mean dispersion $\Delta(s) = \max_k R(\tau_k) - \bar R(s)$, and report the per-decision $R^2$ of the proportional fit through the origin.

\begin{table*}[t]
  \centering
  \caption{Per-$(K, H)$ proportional-fit recovery. $n$ counts pooled decisions across three seeds. The fit is $R^2 \geq 0.98$ on every configuration; $\beta$ varies by less than $3\%$ from the mean across the entire grid. Both $\Delta_c$ and lift grow with $K$ (the order-statistic $\kappa(K)$ effect), and their ratio holds.}
  \label{tab:kh_invariance}
  \footnotesize
  \begin{tabular}{rrrrrrr}
    \toprule
    $K$ & $H$ & $n$ & $\Delta_c$ & sel.\ gain & $\beta_{\mathrm{win}}$ & per-dec $R^2$ \\
    \midrule
    $3$  & $10$ & $211$  & $0.111$ & $0.109$ & $0.983$ & $0.995$ \\
    $3$  & $20$ & $222$  & $0.157$ & $0.155$ & $0.989$ & $0.998$ \\
    $3$  & $40$ & $307$  & $0.206$ & $0.201$ & $0.975$ & $0.992$ \\
    $5$  & $10$ & $246$  & $0.198$ & $0.195$ & $0.986$ & $0.997$ \\
    $5$  & $20$ & $259$  & $0.224$ & $0.222$ & $0.993$ & $0.999$ \\
    $5$  & $40$ & $650$  & $0.223$ & $0.216$ & $0.966$ & $0.989$ \\
    $7$  & $10$ & $288$  & $0.162$ & $0.159$ & $0.979$ & $0.996$ \\
    $7$  & $20$ & $313$  & $0.224$ & $0.216$ & $0.963$ & $0.980$ \\
    $7$  & $40$ & $504$  & $0.259$ & $0.255$ & $0.984$ & $0.996$ \\
    $11$ & $10$ & $207$  & $0.223$ & $0.219$ & $0.983$ & $0.996$ \\
    $11$ & $20$ & $222$  & $0.362$ & $0.351$ & $0.970$ & $0.978$ \\
    $11$ & $40$ & $1{,}085$ & $0.207$ & $0.202$ & $0.975$ & $0.990$ \\
    \midrule
    \multicolumn{2}{l}{\textbf{across 12 configs}} & $4{,}514$ & --- & --- & $\mathbf{0.979 \pm 0.009}$ {\scriptsize [$0.963, 0.993$], CV $0.9\%$} & $\geq 0.98$ \\
    \bottomrule
  \end{tabular}
\end{table*}

\subsection{$K$-Saturation}
\label{app:k_saturation}

Lift saturates by $K{\approx}32$ ($+21.8\% \pm 6.6\%$ SEM, $n{=}5$ seeds); doubling to $K{=}64$ does not separate from $K{=}32$ at our seed budget ($+23.0\% \pm 3.0\%$, $n{=}5$ seeds). The directly-measured selection accuracy $\alpha(K)$ of the value head on the paper-protocol GTrXL dumps is $0.904$ at $K{=}7$ ($n{=}977$ decisions across $12$ runs), $0.870$ at $K{=}32$ ($n{=}338$, $5$ seeds), and $0.844$ at $K{=}64$ ($n{=}346$, $5$ seeds), all under the loose-tie convention in which selecting any branch attaining the maximum return counts as correct, applied consistently across $K$. The tie-strict convention, crediting only a unique argmax, reads $0.344$ at $K{=}7$; the gap is the mass of decisions where several branches tie at the maximum, and the loose reading matches what Lemma~\ref{thm:scoring_bound} uses, since committing any maximizing branch realizes $\Delta(s)$. The decline with $K$ is monotone but modest, $\alpha(7){-}\alpha(64) = 0.060$ while $\kappa(K)$ grows $1.352 \to 2.343$, so the order-statistic gain dominates and lift rises from $K{=}7$ to $K{=}32$ before the $\alpha$ decline bites at $K{=}64$.

\paragraph{Does the tie convention reorder the verifiers?} No. The verifier hierarchy of Table~\ref{tab:slopes} is estimated from lift-versus-dispersion slopes and episode return, neither of which uses $\alpha$. The ordering is stable anyway. On the larger pool of every retained $K{=}7$, $H{=}20$ dump, which includes runs outside the paper protocol ($6{,}675$, $6{,}886$ and $4{,}099$ decisions), loose $\alpha$ is $0.98$, $0.95$, $0.86$ for the value head, Qwen-14B and Qwen-72B and strict $\alpha$ is $0.27$, $0.26$, $0.25$. The level collapses under the strict convention and the ranking does not. These values differ from the $0.904$ and $0.344$ above because the pools differ, chiefly in the share of decisions where every branch ties. Two caveats. Each verifier's dumps come from its own runs, so these are within-verifier diagnostics and not a controlled comparison, and the share of decisions where every branch earns the same return varies from $49\%$ to $65\%$ across those pools. On the $H{=}15$ dumps used off Craftax, the top two swap between conventions, separated by under $0.005$ either way.

\section{Where the Dispersion Comes From: Full Detail}
\label{app:mechanism}

\subsection{Action Choice against Realization: Per-Dump Detail}
\label{app:dispersion_extra}

Companion to Section~\ref{sec:mechanism}. Table~\ref{tab:action_vs_noise} decomposes branch-return variance from the logged decision dumps, and Table~\ref{tab:dispersion_full} disaggregates the dispersion measurement by run and verifier.

\begin{table*}[t]
  \centering
  \caption{\textbf{Dispersion is mostly the environment's under a confident policy.} Decomposition of branch-return variance from the logged decision dumps, into a between-action component (the policy's choice) and a within-action component (the rollout, with the first action held fixed). ``All-same'' is the fraction of decisions at which every branch sampled the identical first action. The $\pi_{\text{strong}}$ row pools $K \in \{3, \ldots, 64\}$; its between-action share at the operating $K{=}7$ is $3.1\%$ on $9{,}008$ decisions. Rows differ in $K$ because the dumps do, and the share is stable in $K$ within each policy.}
  \label{tab:action_vs_noise}
  \footnotesize
  \begin{tabular}{llcccc}
    \toprule
    Environment & Policy & $K$ & Decisions & All-same & Between-action share \\
    \midrule
    Craftax-Medium & $\pi_{\text{strong}}$ (GTrXL)  & $3$--$64$ & $20{,}936$ & $83.7$--$92.3\%$ & $0.9$--$6.2\%$ \\
    Craftax-Medium & $\pi_{\text{weak}}$ (PPO-MLP)  & $3$       & $2{,}252$  & $15.1\%$         & $40.5\%$ \\
    Asterix-MinAtar & $\pi_{\text{weak}}$ (PPO-MLP) & $7$       & $5{,}902$  & $5.7\%$          & $40.0\%$ \\
    DoorKey-8x8 & $\pi_{\text{weak}}$ (PPO-MLP)     & $5$       & $1{,}174$  & $88.9\%$         & $1.6\%$ \\
    \bottomrule
  \end{tabular}
\end{table*}

\begin{table}[!htbp]
  \centering
  \caption{Dispersion bound, per-dump breakdown. Decision dumps from $\pi_{\text{strong}}$ runs at $K{=}7$, $H{=}20$, periodic-50 on Craftax-Symbolic-Medium and Craftax-Easy. ``Sparse'' is the fraction of decisions with $\Delta(s)=0$; $\Delta_c$ is the mean over all decisions and over only informative (non-sparse) decisions.}
  \label{tab:dispersion_full}
  \footnotesize
  \begin{tabular}{lcrrr}
    \toprule
    Oracle & $n$ & Sparse & $\Delta_c$ all & $\Delta_c$ informative \\
    \midrule
    \multicolumn{5}{l}{\emph{Craftax-Symbolic-Medium:}} \\
    Value head   & 66  & $25.8\%$ & $0.74$ & $1.00$ \\
    Qwen-72B-AWQ   & 58  & $25.9\%$ & $0.70$ & $0.95$ \\
    \midrule
    \multicolumn{5}{l}{\emph{Craftax-Easy:}} \\
    Qwen-14B-AWQ (seed 42) & 503  & $75.1\%$ & $0.05$ & $0.19$ \\
    Qwen-14B-AWQ (seed 43) & 976  & $74.3\%$ & $0.04$ & $0.14$ \\
    \bottomrule
  \end{tabular}
\end{table}

\subsection{Common-Random-Numbers Control: Action Choice against Realization}
\label{app:crn_control}

Table~\ref{tab:action_vs_noise} decomposes dispersion observationally, by grouping logged branches on their first action. We report the interventional version of the same question here.

\paragraph{The control.} With \texttt{branching.shared\_env\_rng} enabled, all $K$ branches step the environment from a single key stream at each horizon step while policy sampling stays independent per branch. Sharing the policy keys too would collapse the branches into one trajectory and force $\Delta$ to zero by construction; sharing only the environment's dice means two branches drawing identical actions follow identical trajectories and contribute nothing to $\Delta$, while branches that diverge in action still separate. Residual dispersion is therefore action-driven, and the shortfall against the uncontrolled condition is the environment's. The shared key derives from branch $0$'s, so the draws still vary across decisions and episodes. Design: three seeds at the Craftax-Medium operating point (GTrXL, value head, periodic-50, $K{=}7$, $H{=}20$, $10$ episodes per seed), matched to the uncontrolled condition on checkpoint and configuration. The comparison is distributional and not paired, since enabling the control changes which branches win, so the two conditions visit different states after the first decision.

\begin{table}[!htbp]
  \centering
  \caption{\textbf{Most measured dispersion is the environment's realization, not the policy's action.} Sharing the environment's random draws across branches removes $69\%$ of mean dispersion and roughly doubles the share of decisions with no dispersion at all. The direction holds in every seed. Episode return is reported for direction only: $10$ episodes per seed is far too few to size the effect, and the no-branching baseline for this cell is $32.09$.}
  \label{tab:crn_control}
  \footnotesize
  \begin{tabular}{lcccc}
    \toprule
    & \multicolumn{2}{c}{Independent dice} & \multicolumn{2}{c}{Shared dice (CRN)} \\
    \cmidrule(lr){2-3} \cmidrule(lr){4-5}
    Seed & Decisions & Mean $\Delta$ & Decisions & Mean $\Delta$ \\
    \midrule
    $42$ & $66$  & $0.777$ & $78$  & $0.092$ \\
    $43$ & $73$  & $0.514$ & $57$  & $0.405$ \\
    $44$ & $155$ & $0.290$ & $147$ & $0.066$ \\
    \midrule
    Pooled & $294$ & $0.455$ & $282$ & $0.142$ \\
    \midrule
    \multicolumn{5}{l}{\emph{Pooled contrast:} $\Delta$ difference $-0.313$, SE $0.072$, $95\%$ CI $[-0.453, -0.173]$, $z = 4.37$} \\
    \multicolumn{5}{l}{\emph{Share of decisions with} $\Delta = 0$: $58.5\%$ independent against $84.8\%$ CRN} \\
    \multicolumn{5}{l}{\emph{Episode return:} $36.00$ independent against $31.53$ CRN, baseline $32.09$} \\
    \bottomrule
  \end{tabular}
\end{table}

\paragraph{Reading the two numbers together.} The control leaves $31\%$ of mean dispersion standing while Table~\ref{tab:action_vs_noise} attributes $3.1\%$ of branch-return variance to the first action. They count different things: policy sampling stays free at all $20$ horizon steps under the control, so its residual absorbs action divergence anywhere in the window, while $3.1\%$ isolates the first action alone. Together they bracket the action channel, small at the committed decision and modest across the window, and in neither reading the majority of what branching harvests.

\paragraph{Scope.} The control supports the claim that fork-and-resume banks realizations rather than actions. It does not establish an episode-return effect: the return column of Table~\ref{tab:crn_control} rests on $30$ episodes per condition against a per-seed spread near $8$. An operator who cannot resume, and must re-branch from the winning first action, is closer to the CRN condition than to ours, and what this says about that operator is that the residual action channel is real but small.

\subsection{The Classical Rollout Algorithm on the Same Rollouts}
\label{app:rollout_contrast}

The rollout algorithm~\citep{tesauro1996online, bertsekas1997rollout} branches as we do but averages the branches sharing a first action and commits the \emph{action}. It can therefore only move return between action groups, so on the logged branches its within-window gain is $\max_a \operatorname{mean}(R \mid a) - \bar R(s)$ against our $\max_k R(\tau_k) - \bar R(s)$. Both read the same $K$ rollouts, so no new simulation is needed and the pool matches Table~\ref{tab:action_vs_noise} decision for decision. The contrast is within-window only: the rollout algorithm resumes from a different state, so a matched episode comparison would require new runs and we do not claim one.

Where every branch draws the same first action there is a single group and the rollout algorithm's gain is zero by construction. Table~\ref{tab:rollout_contrast} shows that this is the common case under a confident policy, and that the captured fraction tracks the between-action variance share of Table~\ref{tab:action_vs_noise}: what the rollout algorithm recovers follows the action channel, which is the channel our decomposition finds nearly empty under $\pi_{\text{strong}}$. The relation is monotone rather than proportional, since two cells at nearly the same between-action share ($40.5\%$ and $40.0\%$) read $0.697$ and $0.496$. The live ratio also falls in $K$ ($0.465$, $0.177$, $0.153$ at $K{=}7$, $16$, $64$), as it must: branch-commit's gain grows with the expected maximum $\kappa(K)$ while the action-group means concentrate as each group gains branches, shrinking the numerator as the denominator grows. Intervals are $10{,}000$-iteration bootstrap resamples over dumps, weighted by decisions.

\begin{table*}[t]
  \centering
  \caption{Within-window gain of the classical rollout algorithm against branch-commit, on identical logged rollouts. ``Degenerate'' is the share of decisions at which every branch drew the same first action, where the rollout algorithm has nothing to select over. ``Ratio'' is rollout gain over branch-commit gain, with a bootstrap $95\%$ interval over dumps; ``live'' restricts to non-degenerate decisions. ``Between-action'' repeats the variance share from Table~\ref{tab:action_vs_noise}. $^{\dagger}$DoorKey-8x8 is a zero-lift environment (Table~\ref{tab:cross_env_dispersion}), so we read its ratio as uninformative despite the tight interval.}
  \label{tab:rollout_contrast}
  \scriptsize
  \begin{tabular}{llcccccc}
    \toprule
    Environment & Policy & $K$ & Degenerate & Between-action & Ratio & $95\%$ CI & Ratio (live) \\
    \midrule
    Craftax-Medium  & $\pi_{\text{strong}}$ & $7$ & $84.9\%$ & $3.1\%$  & $0.106$ & $[0.064, 0.165]$ & $0.465$ \\
    Craftax-Medium  & $\pi_{\text{strong}}$ & $16$ & $85.5\%$ & $2.9\%$ & $0.031$ & $[0.024, 0.040]$ & $0.177$ \\
    Craftax-Medium  & $\pi_{\text{strong}}$ & $64$ & $84.2\%$ & $2.4\%$ & $0.035$ & $[0.026, 0.042]$ & $0.153$ \\
    Craftax-Medium  & $\pi_{\text{weak}}$   & $3$ & $15.1\%$ & $40.5\%$ & $0.697$ & $[0.666, 0.725]$ & $0.814$ \\
    Asterix-MinAtar & $\pi_{\text{weak}}$   & $7$ & $5.7\%$  & $40.0\%$ & $0.496$ & $[0.487, 0.504]$ & $0.523$ \\
    DoorKey-8x8$^{\dagger}$ & $\pi_{\text{weak}}$ & $5$ & $88.9\%$ & $1.6\%$  & $0.029$ & $[0.020, 0.042]$ & $0.419$ \\
    \bottomrule
  \end{tabular}
\end{table*}

\paragraph{Scope.} The contrast says what the rollout algorithm can extract from the candidates branching already generates, not which method an operator should run at a matched simulator budget. The rollout algorithm spends its budget differently, re-branching from the committed action rather than resuming, and comparing the two as deployed policies needs the matched-budget runs we do not have. The claim here is narrower and sufficient for the mechanism: committing the action discards most of what the $K$ rollouts contain exactly when the policy is confident.

\subsection{Does the Window Measure Miss Exploitable Spread?}
\label{app:latent_dispersion}

Section~\ref{sec:discussion} identifies environments with delayed rewards not visible within the branch window as an important direction, and our grid contains no such environment. A related test needs no new environment: the relevant \emph{decisions} are already inside the one we ran. Dispersion is measured on the realized $H$-step return $R$, so on decisions where every branch earns an identical $R$ it is exactly zero by construction. If those branches nonetheless differ in the bootstrap $\hat V(s_H)$, the window measure understates the spread a value-head verifier can act on, which is the delayed-payoff limitation in miniature.

Pooling the $K{=}7$, $H{=}20$ decision dumps across verifier classes ($18{,}031$ decisions), $60.8\%$ of decisions have $\Delta_R = 0$. On $97.5\%$ of those the branches still differ in $\hat V(s_H)$, with mean $\Delta_{\hat V} = 0.135$, and the deployed verifier selects on that residual rather than at random, placing its chosen branch at the $0.868$ percentile of $\hat V$ within its own decision. For the value head the percentile is $1.000$, which is the expected degenerate case: once $R$ ties across branches, the bootstrapped score reduces to $\hat V(s_H)$ alone, so its argmax is the $\hat V$ argmax by definition. For contrast, on the $7{,}075$ decisions where $\Delta_R > 0$ the mean is $\Delta_R = 0.690$ with $\Delta_{\hat V} = 0.543$, so the residual spread on flat decisions is roughly a quarter of what is available where the window is informative.

So the window measure understates the exploitable spread, and the understatement is bounded and does not dominate. The dumps cannot show whether acting on the residual raises episode return, since they carry no counterfactual for the branch not taken. $\Delta$ as we measure it is therefore the right bound for a verifier reading realized within-window return and a slightly conservative one for a verifier that also reads a critic.

\subsection{Tail Behaviour of Branching}
\label{app:tail_risk}

Companion to Section~\ref{sec:discussion}. Every headline number here is a mean episode return, and branching commits an $H$-step trajectory selected for a high \emph{maximum}, which should change the spread of outcomes as well as their centre. We pull per-episode returns from the canonical GTrXL Craftax-Medium checkpoint and separate arms by each run's own logged configuration and not by job name, since the trigger ablations inherit \texttt{branch\_frequency}~$=50$ while firing on a different signal and would otherwise pool into the uniform arm. The branched arm is restricted to the value-head verifier.

\begin{table}[!htbp]
  \centering
  \caption{Per-episode return distributions on Craftax-Medium, canonical checkpoint. The branched arm is uniform periodic-$50$ under the value head. Rates are the fraction of episodes returning less than the stated fraction of the no-branching mean ($32.72$ on this pool).}
  \label{tab:tail_risk}
  \footnotesize
  \resizebox{\columnwidth}{!}{\begin{tabular}{lccccccc}
    \toprule
    Arm & Runs & Episodes & Mean & SD & $<50\%$ & $<75\%$ & $<100\%$ \\
    \midrule
    No branching            & $38$  & $870$  & $32.72$ & $8.08$ & $3.0\%$ & $18.2\%$ & $34.5\%$ \\
    Periodic-50 (VF)        & $113$ & $2656$ & $34.38$ & $9.49$ & $6.1\%$ & $17.5\%$ & $27.5\%$ \\
    \bottomrule
  \end{tabular}}
\end{table}

Two things move in opposite directions. Branching shifts the distribution up, and the rates of moderate underperformance fall: episodes below three quarters of the no-branching mean go from $18.2\%$ to $17.5\%$, and below the mean itself from $34.5\%$ to $27.5\%$. But the standard deviation rises from $8.08$ to $9.49$, and the catastrophic rate roughly doubles, from $3.0\%$ to $6.1\%$. That difference is resolved rather than noise: $+3.1$ points with a $95\%$ interval of $[+1.7, +4.6]$ under a two-proportion normal approximation ($z = 4.2$).

The two pools differ in run composition and per-run episode counts, so this is a distributional measurement and not a matched contrast. The direction is what the mechanism predicts: an operator whose loss is asymmetric in bad episodes should not read a mean improvement as a uniform one, and should price the tail on their own deployment.

\section{Verifier Comparison: Full Detail}
\label{app:ablations}

\subsection{Operating-Point Comparison and Cost}
\label{app:operating_point}

Figure~\ref{fig:k7_tie} gives the confidence intervals behind the value-head and GPT-4o comparison of Section~\ref{sec:investigation}, and Table~\ref{tab:asterix_tie} repeats the comparison on Asterix-MinAtar.

\begin{figure}[!htbp]
  \centering
  \includegraphics[width=\columnwidth]{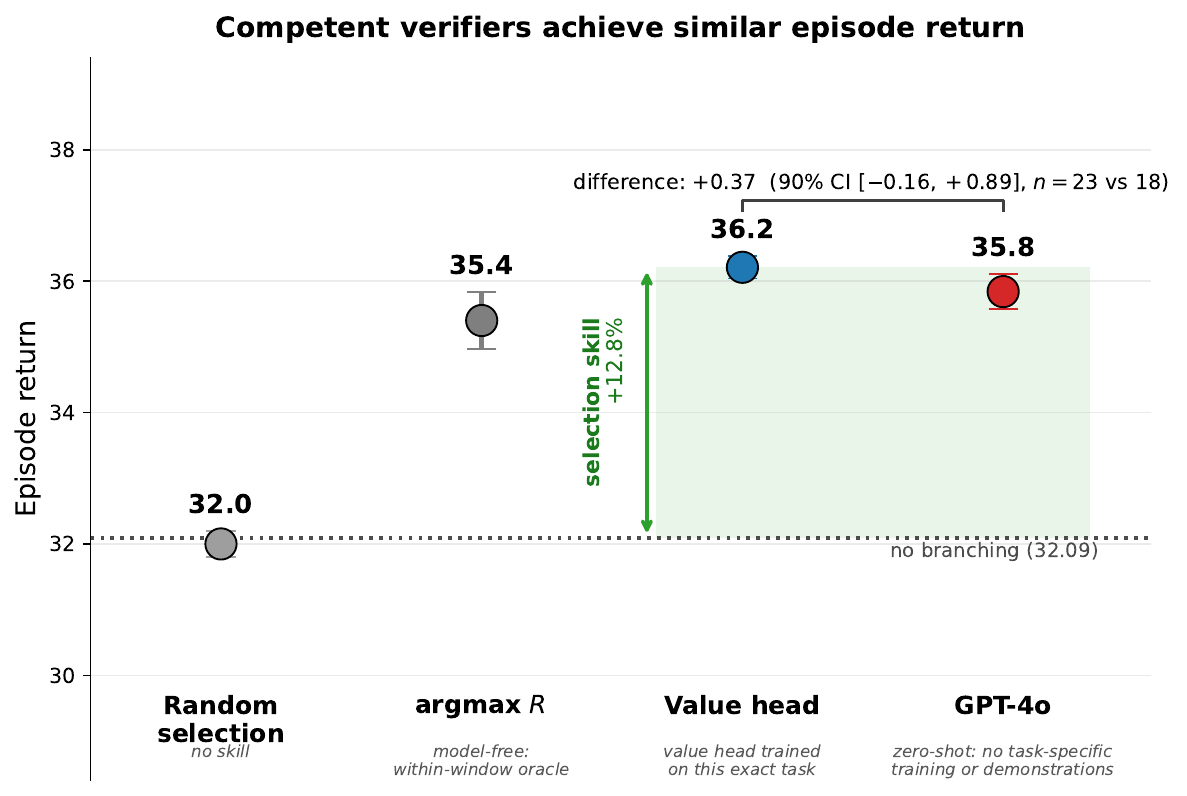}
  \caption{\textbf{Value head and zero-shot GPT-4o at the operating budget.} Episode return at the operating budget ($K{=}7$, periodic-50, $H{=}20$, Craftax-Medium), mean $\pm$ Student-$t$ $95\%$ CI. The trained value head ($n{=}23$ runs) and zero-shot GPT-4o ($n{=}18$ runs) differ by $+0.37$ ($90\%$ CI $[-0.16, +0.89]$; the $95\%$ interval is $[-0.26, +0.99]$), whose far end is under a third of the branching gain on either baseline convention (Section~\ref{sec:methodology}); both clear random selection on the no-branching baseline.}
  \Description{Bar chart of mean episode return with 95 percent confidence intervals on Craftax-Medium at K equals 7. The trained value head and zero-shot GPT-4o have nearly equal returns with overlapping intervals, and both lie above random selection and the no-branching baseline.}
  \label{fig:k7_tie}
\end{figure}

\begin{table}[!htbp]
  \centering
  \caption{Second-environment operating-point tie on Asterix-MinAtar ($\pi_{\text{weak}}$ PPO-MLP, $K{=}7$, $H{=}15$, periodic-15, $3$ seeds). All four language-model verifiers land within seed noise of the trained value head: Welch $|t| < 1$ for every VF--LLM comparison, so none separates from the value head. The no-branching row is the paired baseline of Table~\ref{tab:cross_env_dispersion}. VF's seed-level spread is wide ($\pm 2.13$), so even the lower-scoring Qwen verifiers do not separate, and three seeds cannot bound the gap as Section~\ref{sec:investigation} does on Craftax-Medium.}
  \label{tab:asterix_tie}
  \footnotesize
  \begin{tabular}{lc}
    \toprule
    Verifier & Episode return \\
    \midrule
    Value head        & $14.30$ \\
    GPT-4o-mini         & $14.58$ \\
    GPT-4o              & $13.58$ \\
    Qwen-2.5-14B-AWQ    & $11.74$ \\
    Qwen-2.5-72B-AWQ    & $10.91$ \\
    \midrule
    No branching (paired eval) & $1.37 \pm 0.08$ \\
    \bottomrule
  \end{tabular}
\end{table}

\paragraph{Cost asymmetry.} The bootstrapped value verifier is sub-millisecond per call (a single forward pass through the existing critic head, already shipped with the policy); GPT-4o costs $\sim\$2$ per 10-episode evaluation at periodic-50 over the OpenAI API; GPT-4o-mini reduces that to $\sim\$0.20$; the open-weight Qwen verifiers are GPU-resident with no API cost but require co-locating a 14B--72B model alongside the policy. Wall-clock numbers vary widely on our shared cluster and are not reported as benchmarks.

\subsection{Where the LLM--VF Per-Episode Tie Comes From: Per-Achievement Decomposition}
\label{app:per_achievement}

The $K{=}7$ episode-return tie between GPT-4o and the value head ($35.84 \pm 2.13$ against $36.21 \pm 1.88$, Section~\ref{sec:investigation}) absorbs a compositional difference in \emph{which} branches each verifier picks. Decomposing per-episode return into per-achievement contributions ($0/1$ flags per episode, three seeds per condition) suggests the mean tie averages opposing signals: the LLM concentrates wins on rare long-horizon achievements and trails VF on common short-horizon ones. No individual achievement survives multiplicity correction, so this is directional evidence and not a resolved per-achievement result.
\begin{table*}[t]
  \centering
  \caption{Per-achievement rates on Craftax-Symbolic-Medium $\pi_{\text{strong}}$ (GTrXL) at the main protocol ($K{=}7$, $H{=}20$, periodic-50, severity 0). Rates are mean per-episode achievement-flag values across seeds. \emph{Baseline} is no-branching; \emph{VF} is value-bootstrap at $K{=}7$; \emph{LLM} is the API-model selector at $K{=}7$ (Qwen-14B-AWQ gives the same direction with a smaller gap). Rows are ordered by $|\text{LLM}-\text{VF}|$, so the top rows are selected on effect size and their gaps are correspondingly optimistic; the paragraph below attaches episode-level uncertainty, which none of them survives.}
  \label{tab:per_achievement}
  \footnotesize
  \setlength{\tabcolsep}{4pt}
  \begin{tabular}{lcccrrr}
    \toprule
    Achievement & Base & VF & LLM & VF$-$Base & LLM$-$Base & LLM$-$VF \\
    \midrule
    \multicolumn{7}{l}{\emph{Where LLM wins (long-horizon / rare):}} \\
    collect\_diamond     & $0.025$ & $0.050$ & $\mathbf{0.233}$ & $+0.025$ & $+0.208$ & $\mathbf{+0.183}$ \\
    make\_iron\_sword    & $0.050$ & $0.100$ & $\mathbf{0.167}$ & $+0.050$ & $+0.117$ & $+0.067$ \\
    make\_iron\_pickaxe  & $0.000$ & $0.117$ & $\mathbf{0.167}$ & $+0.117$ & $+0.167$ & $+0.050$ \\
    make\_stone\_sword   & $0.775$ & $0.733$ & $0.833$ & $-0.042$ & $+0.058$ & $+0.100$ \\
    collect\_iron        & $0.675$ & $0.567$ & $0.667$ & $-0.108$ & $-0.008$ & $+0.100$ \\
    \midrule
    \multicolumn{7}{l}{\emph{Where VF wins (short-horizon / dense):}} \\
    attack\_cow          & $0.775$ & $\mathbf{0.833}$ & $0.733$ & $+0.058$ & $-0.042$ & $-0.100$ \\
    defeat\_skeleton     & $0.500$ & $\mathbf{0.450}$ & $0.400$ & $-0.050$ & $-0.100$ & $-0.050$ \\
    \midrule
    \multicolumn{7}{l}{\emph{Where both verifiers tie or saturate:}} \\
    collect\_wood, mine\_stone, etc. & near-sat. & near-sat. & near-sat. & $\sim 0$ & $\sim 0$ & $\sim 0$ \\
    \bottomrule
  \end{tabular}
\end{table*}

\paragraph{Uncertainty and multiplicity.} The rates above are seed-level means. Recovering per-episode achievement flags from the logged running means (every de-cumulated value lies within $10^{-6}$ of $0$ or $1$) lets us attach episode-level uncertainty and correct across the $18$ achievements present in both arms. No achievement separates the verifiers. For collect\_diamond, the largest cell, the gap is $+0.167$ ($7/30$ against $2/30$) with a bootstrap $95\%$ CI of $[+0.000, +0.333]$, uncorrected $p = 0.071$, Holm-corrected $p = 1.000$; every other achievement is wider and also reads Holm $p = 1.000$. Substituting Qwen-2.5-14B-AWQ gives the same verdict (collect\_diamond $0.175$ against VF's $0.067$, Holm $p = 1.000$). Two provenance notes: the rates above match the API-model (\texttt{selector:llm}) group and not the Qwen-14B group, and the VF arm recoverable from the local run cache covers $30$ of the $60$ episodes behind the $0.050$ figure, reading $0.067$. Neither changes the direction of any row.

The collect\_diamond row shows the direction of the effect, and the Holm-corrected test above shows that it is not resolved: the policy already collects diamond at $2.5\%$ of episodes at baseline; the value head at $K{=}7$ raises this to $5.0\%$ ($2\times$ baseline); the LLM raises it to $23.3\%$ ($\sim 9\times$ baseline, $\sim 5\times$ VF). Diamond requires a chained tool progression (wood pickaxe $\to$ mine stone $\to$ make stone pickaxe $\to$ mine iron $\to$ make iron pickaxe $\to$ mine diamond), and each step's contribution to per-episode return is delayed and only loosely correlated with the critic's terminal value $\hat V(s_H)$ at any single branching decision. The LLM, ranking branches on a reward-derived axis (within-decision $\bar\rho(\text{LLM}, R_H) = +0.50$ across providers, Appendix~\ref{app:spearman_per_llm}), recognizes which branch begins or extends the chain, even when the critic's $\hat V(s_H)$ across branches is essentially flat ($\bar\rho(\text{LLM}, \hat V(s_H)) \approx 0$). On short-horizon dense decisions like \emph{attack\_cow}, the critic's $\hat V(s_H)$ already encodes the right ranking and the value head matches reward closely. The per-episode mean tie is what aggregates these opposing per-achievement signals.

\subsection{Within-Decision Spearman Audit: Per-LLM Breakdown}
\label{app:spearman_per_llm}

LLM verifiers rank branches by realized $H$-step reward and not by the critic's terminal value: within a decision, every LLM's scores correlate with $R_H$ at $\bar\rho \approx 0.5$ and with $\hat V(s_H)$ at $\bar\rho \approx 0$ (Table~\ref{tab:spearman_per_llm}). Where an LLM and the value head disagree, the LLM's pick trails by an amount that grows with dispersion (Table~\ref{tab:tertile_residuals}, Figure~\ref{fig:cross_llm_regime_law}). For each LLM verifier with main-protocol GTrXL coverage on Craftax-Symbolic-Medium, we compute, per branched decision, the rank correlation between the LLM's branch scores and (a) the critic's terminal value $\hat V(s_H^{(k)})$ across $K$ branches, and (b) the realized $H$-step branch reward $R_H^{(k)}$. Decisions are then averaged within each LLM. The audit reads the per-step decision logs that record both LLM scores and per-branch terminal-value forecasts.

\begin{table}[!htbp]
  \centering
  \caption{Per-LLM within-decision Spearman audit on Craftax-Symbolic-Medium with $\pi_{\text{strong}}$ (GTrXL), $K{=}7$, $H{=}20$, periodic-50. $n_{\text{dec}}$ counts informative branched decisions (with non-degenerate score variance). The V-blind reward-correct pattern holds for every LLM tested: every value of $\bar\rho(\text{LLM}, \hat V(s_H))$ is indistinguishable from zero, while every value of $\bar\rho(\text{LLM}, R_H)$ is in $[+0.49, +0.55]$. The pattern is consistent across two providers (OpenAI, Alibaba), two model families (GPT, Qwen), and two model sizes (14B, 72B).}
  \label{tab:spearman_per_llm}
  \footnotesize
  \begin{tabular}{lcccc}
    \toprule
    LLM verifier & runs & $n_{\text{dec}}$ & $\bar\rho(\text{LLM}, \hat V(s_H))$ & $\bar\rho(\text{LLM}, R_H)$ \\
    \midrule
    GPT-4o            & $27$ & $4{,}958$ & $+0.009$ & $+0.546$ \\
    Qwen-2.5-14B-AWQ  & $24$ & $2{,}054$ & $+0.005$ & $+0.504$ \\
    Qwen-2.5-72B-AWQ  & $11$ & $2{,}518$ & $-0.001$ & $+0.499$ \\
    \bottomrule
  \end{tabular}
\end{table}

\begin{table*}[t]
  \centering
  \caption{\textbf{Per-LLM tertile residuals, raw and normalized.} Mean signed verifier-pick gap $\delta_{\text{pick}} = R(\tau_{\text{LLM-pick}}) - R(\tau_{\text{VF-pick}})$ in $H{=}20$ step reward units, conditional on the LLM and value-head verifiers disagreeing on the top-1 branch, partitioned by within-decision dispersion tertile (cutoffs $0.20$, $0.91$, quantiles over all informative decisions). $n$ counts informative disagreement events per cell, over $109$ main-protocol runs. The normalized rows divide each gap by the dispersion available at its decision.}
  \label{tab:tertile_residuals}
  \footnotesize
  \begin{tabular}{lcccc}
    \toprule
    LLM verifier & low $\Delta\,(n)$ & mid $\Delta\,(n)$ & high $\Delta\,(n)$ & all $\Delta\,(n)$ \\
    \midrule
    \multicolumn{5}{l}{\emph{Raw} $\delta_{\text{pick}}$:} \\
    GPT-4o          & $-0.033\,(540)$ & $-0.466\,(649)$ & $-1.749\,(640)$ & $-0.787\,(1{,}829)$ \\
    GPT-4o-mini     & $+0.007\,(663)$ & $-0.192\,(663)$ & $-1.712\,(637)$ & $-0.618\,(1{,}963)$ \\
    Qwen-2.5-14B-AWQ & $-0.025\,(489)$ & $-0.303\,(399)$ & $-2.007\,(421)$ & $-0.747\,(1{,}309)$ \\
    Qwen-2.5-72B-AWQ & $-0.008\,(568)$ & $-0.345\,(629)$ & $-2.069\,(588)$ & $-0.806\,(1{,}785)$ \\
    \midrule
    Pooled          & $-0.013\,(2{,}260)$ & $-0.328\,(2{,}340)$ & $-1.869\,(2{,}286)$ & $-0.736\,(6{,}886)$ \\
    \midrule
    \multicolumn{5}{l}{\emph{Normalized} $\delta_{\text{pick}}/\Delta$, pooled:} \\
    Mean            & $+0.002$ & $-0.619$ & $-0.748$ & $-0.458$ \\
    $95\%$ CI       & $[-0.107, +0.119]$ & $[-0.678, -0.563]$ & $[-0.780, -0.715]$ & \\
    Mean $\Delta$ in bin & $0.077$ & $0.580$ & $2.538$ & \\
    \bottomrule
  \end{tabular}
\end{table*}

\begin{figure}[!htbp]
  \centering
  \includegraphics[width=\columnwidth]{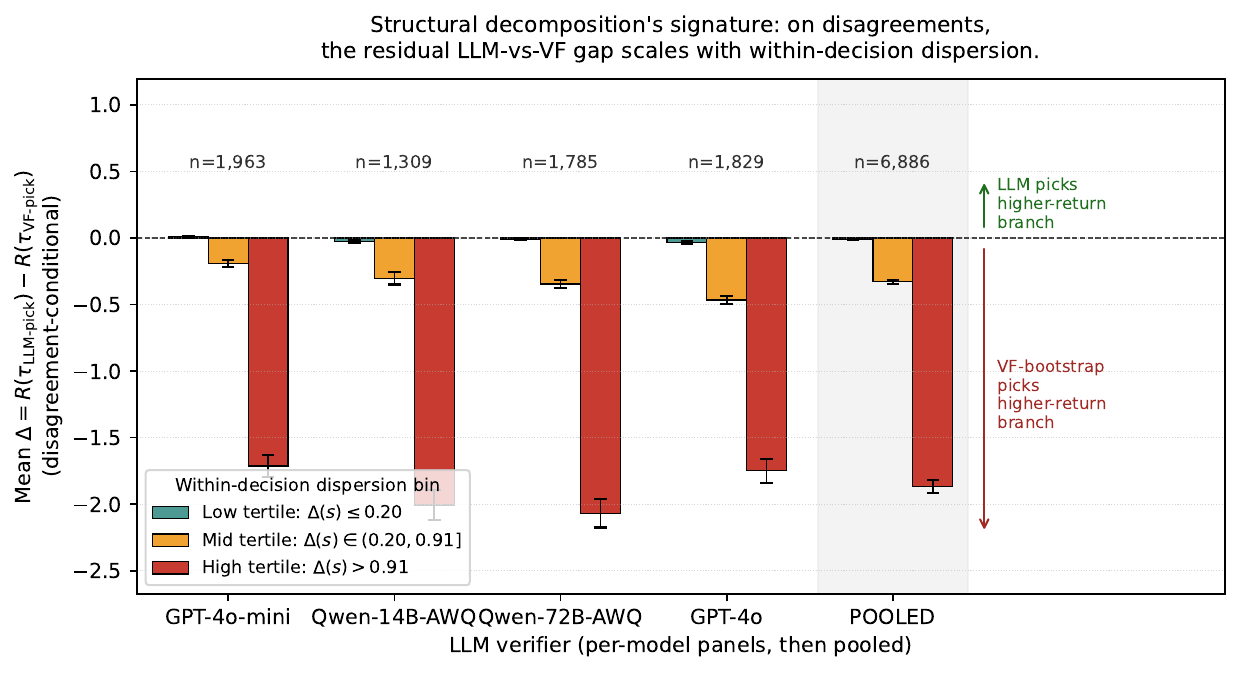}
  \caption{\textbf{Raw verifier-pick gap by dispersion tertile.} Mean signed gap $\delta_{\text{pick}} = R(\tau_{\text{LLM-pick}}) - R(\tau_{\text{VF-pick}})$ on decisions where the LLM and the value head disagree, by within-decision dispersion tertile (cutoffs $0.20$ and $0.91$ in $H$-step reward units). Bars are pooled means $\pm$ standard error over $n{=}6{,}886$ disagreements, the raw rows of Table~\ref{tab:tertile_residuals}. GPT-4o-mini is on the MLP policy, the other panels on GTrXL.}
  \Description{Panels of bar charts, one per language-model verifier, showing the mean reward gap between the LLM's pick and the value head's pick on disagreements, split into low, mid, and high dispersion tertiles with standard-error bars. The gap is near zero in the low tertile and grows more negative as dispersion increases.}
  \label{fig:cross_llm_regime_law}
\end{figure}

The normalized rows of Table~\ref{tab:tertile_residuals} separate the dispersion trend from its mechanical part. A larger $\Delta$ scales the stakes of any disagreement, so raw gaps grow with dispersion by construction. After dividing by $\Delta$, the low-tertile gap stays at zero ($+0.002$, CI $[-0.107, +0.119]$), and the mid and high tertiles are similar ($-0.619$ and $-0.748$), so most of the raw growth from mid to high is mechanical. Disagreements are a minority of decisions; over all decisions the LLM verifiers come close to the value head (Section~\ref{sec:investigation}).

GPT-4o-mini is omitted from Table~\ref{tab:spearman_per_llm} because its decision dumps at this protocol do not record the per-branch terminal-value forecasts the audit correlates against; its episode-return result is reported in Section~\ref{sec:investigation}. Appendix~\ref{app:llm_failure_modes} gives the case-level view of the same pattern.

\subsection{LLM Verifier Failure Modes: Case-Level Categorization}
\label{app:llm_failure_modes}

Case-level companion to the within-decision Spearman pattern of Appendix~\ref{app:spearman_per_llm}. We categorize high-dispersion cases where the LLM picked a strictly worse branch: LLM and the value head disagreed, the LLM's pick had strictly lower realized reward, and within-decision dispersion sat in the top tertile. DoorKey is excluded, since XLand-MiniGrid branches carry no semantic descriptions. That yields $48$ cases ($28$ GPT-4o, $20$ Qwen-14B-AWQ), sent to GPT-4o-2024-08-06 with a fixed five-bucket taxonomy at temperature $0$, given the captions, the LLM's reasoning, all branch rewards, and both pick indices.

The categories are: under-weighting deferred achievements (long\_horizon), over-weighting survival stats (stat\_priority), misreading position (spatial\_nav), misjudging crafting prerequisites (craft\_chain), and a residual bucket. The first two account for $39.6\%$ and $25.0\%$ of cases, $64.6\%$ together, the short-horizon bias the Spearman pattern names structurally; the split is consistent across both failed models ($60.7\%$ for GPT-4o cases, $70.0\%$ for Qwen).

\paragraph{Self-evaluation bias check.} The categorizer is GPT-4o and $58\%$ of the failures are GPT-4o's own, so anti-self-criticism bias would push its failures into the residual bucket. The opposite obtained ($17.9\%$ residual for GPT-4o cases against $5.0\%$ for Qwen-14B). At $n{=}48$ this supports a qualitative reading only.

\subsection{Scrambled-Caption Control}
\label{app:negative_controls}

LLM lift collapses when the captions lose their semantic content. We shuffle the natural-language captions across decisions before submitting to Qwen-14B-AWQ for scoring (3 seeds, paired with the canonical no-branching baseline). The Qwen-14B lift drops from $+7.0\%$ (canonical captions) to $-4.9\%$ (scrambled), consistent with the gain being semantic-content-driven rather than a branching-mechanics or prompt-template effect.

\paragraph{Cross-provider replication on GPT-4o-mini.} The scrambled-caption collapse is not specific to the Qwen family. Running the same protocol (GTrXL, $K{=}7$, $H{=}20$, periodic-50, $10$ episodes per seed) with GPT-4o-mini as the verifier, scrambled-caption episode return is $32.85$ (mean of seeds $42$ and $43$; a third seed was attempted but failed in execution), versus the canonical GPT-4o-mini verifier mean of $35.93$ (seed $42$ aggregate across seven independent runs of $50$ episodes each) and the matched seed-$42$ no-branching baseline of $32.82$ ($50$ episodes). The implied lift drops from $+9.5\%$ (canonical) to $+0.1\%$ (scrambled), reproducing on a frontier API provider and a different model family the same collapse Qwen-14B-AWQ shows on a local-AWQ inference path. Across both providers and model families, the LLM verifier's gain depends on the caption content and is not an artifact of one inference pipeline.

\subsection{Return-Field Ablation on the Caption}
\label{app:return_field_ablation}

The scrambled-caption control of Appendix~\ref{app:negative_controls} displaces the whole caption, so it cannot separate the model reading trajectory content from the model reading the one numeric field the renderer surfaces, the branch's realized $H$-step return. We re-render with that field suppressed and re-run the verifier.

\paragraph{Protocol.} The renderer drops the per-branch return and terminal-value line, the per-step reward lines, and the compact-format reward and value tokens; states, inventory, equipment, position, achievements, threats, action summaries, and event summaries are emitted unchanged. Verified byte-identical to the canonical renderer when the ablation is off, and verified on a live decision that the ablated prompt contains no occurrence of the return, reward, or value fields. Everything else matches the canonical protocol exactly (GTrXL policy, Craftax-Symbolic-Medium, Qwen-2.5-14B-AWQ at temperature $0$, $K{=}7$, $H{=}20$, periodic-$50$, $10$ episodes per seed, seeds $42$--$44$, common random numbers matched to the canonical runs).

\paragraph{Comparison pool.} The canonical Qwen-14B runs on this checkpoint are not homogeneous: some use the optimized prompt template, some ran through the OOD harness with an explicit environment id, some run $30$ episodes. We compare only against runs matching the ablated configuration on every axis (ten episodes, in-domain, optimized prompts off), leaving four canonical runs per seed.

\paragraph{Result.} We run two paired comparisons and report the prospective one first. At fresh seeds $45$--$47$ we launched ablated and control arms from the same script with the flag as the only difference. The ablated arm returns $35.60$, $35.60$, $35.10$ (mean $35.43$) against the control's $34.40$, $34.50$, $32.50$ (mean $33.80$), giving per-seed differences $+1.20$, $+1.10$, $+2.60$ and a mean of $+1.63$ (SEM $0.48$). The second comparison is the weaker of the two, since it reuses existing canonical runs at seeds $42$--$44$ rather than arms launched together: ablated returns $36.20$, $31.10$, $31.90$ against configuration-matched canonical means $34.60$, $34.22$, $30.98$, giving per-seed differences $+1.59$, $-3.13$, $+0.92$ and a mean of $-0.20$ (SEM $1.47$). Pooling all six paired seeds gives $+0.72$ (SEM $0.81$), and that pooled figure is the one we quote. Withholding the realized-return field costs nothing measurable, and the point estimate runs slightly in its favor.

\paragraph{Reading.} Were the model simply ranking the printed number, suppressing it should have driven selection toward chance, which scores $32.00$ and is indistinguishable from not branching. Instead the ablated arm holds its position, pooling to $34.25$ over six seeds. The language model is therefore reading trajectory content, and the realized-return field is not what carries its ranking. The direction of the point estimate, slightly favoring the ablated captions, is consistent with the failure-mode taxonomy of Appendix~\ref{app:llm_failure_modes}, where the dominant error is over-weighting short-horizon signal: removing an explicitly myopic field would if anything reduce that pull. We do not claim that effect, which three seeds cannot resolve; the claim is the null one, that the field is not load-bearing.

\section{Candidate Generation and Dispersion Gating: Full Detail}
\label{app:triggering}

\subsection{Cross-Mechanism Proposer Ablation}
\label{app:b2_proposer}

\begin{figure*}[t]
  \centering
  \definecolor{pxInk}   {HTML}{1f2933}
\definecolor{pxSlate} {HTML}{546069}
\definecolor{pxMute}  {HTML}{aab2b8}
\definecolor{pxRule}  {HTML}{d4dade}
\definecolor{pxOwn}   {HTML}{0d6e7a}

\begin{tikzpicture}[x=1mm, y=1mm,
    >={Stealth[length=1.8mm,width=1.3mm]},
    every node/.style={font=\sffamily, inner sep=0pt, outer sep=0pt},
    barEng/.style={fill=pxMute, draw=pxSlate, line width=0.2pt},
    barOwn/.style={fill=pxOwn,  draw=pxOwn,   line width=0.2pt},
    ax/.style={draw=pxSlate, line width=0.4pt},
    tick/.style={font=\sffamily\tiny, text=pxSlate},
    vlab/.style={font=\sffamily\scriptsize, text=pxInk},
]
\def\bw{3} %

\def\cA{8}\def\cB{17}\def\cC{26}\def\cD{35}\def\cE{44}\def\cF{53}
\draw[pxRule, line width=0.3pt, dashed] (4,9.18) -- (56,9.18);
\fill[barEng] (\cA-\bw,0) rectangle (\cA+\bw,6.84);
\fill[barEng] (\cB-\bw,0) rectangle (\cB+\bw,7.92);
\fill[barOwn] (\cC-\bw,0) rectangle (\cC+\bw,9.18);
\fill[barEng] (\cD-\bw,0) rectangle (\cD+\bw,9.36);
\fill[barEng] (\cE-\bw,0) rectangle (\cE+\bw,11.70);
\fill[barEng] (\cF-\bw,0) rectangle (\cF+\bw,15.48);
\foreach \cx/\h/\v in {\cA/6.84/0.38, \cB/7.92/0.44, \cC/9.18/0.51, \cD/9.36/0.52, \cE/11.70/0.65, \cF/15.48/0.86}
   { \node[tick, anchor=south] at (\cx, \h+0.5) {\v}; }
\draw[ax] (4,0) -- (56,0);
\draw[ax] (4,0) -- (4,18.5);
\foreach \v/\yy in {0/0, 0.5/9, 1.0/18} { \node[tick, anchor=east] at (3, \yy) {\v}; }
\node[vlab, rotate=90, anchor=south] at (-3, 9) {manufactured $\mathbb{E}[\Delta(s)]$};
\draw[->, pxSlate, line width=0.5pt] (\cC+2, 16.6) -- (\cF-2, 16.6);
\node[font=\sffamily\tiny\itshape, text=pxSlate, anchor=south] at (39, 16.9) {engineered ${\sim}2\times$};
\foreach \cx/\name in {%
    \cA/{Tempered $T{=}2$}, \cB/{LLM-prefix $\ell{=}3$},
    \cC/{Stochastic $\star$}, \cD/{Top-$K$ distinct},
    \cE/{LLM-prefix $\ell{=}1$}, \cF/{Counterfactual}}
  { \node[font=\sffamily\tiny, text=pxInk, rotate=32, anchor=north east] at (\cx+1.3, -1) {\name}; }

\def\dA{73}\def\dB{82}\def\dC{91}\def\dD{100}\def\dE{109}\def\dF{118}
\draw[pxOwn, line width=0.45pt, dashed] (69,15.75) -- (121,15.75);
\fill[barEng] (\dA-\bw,0) rectangle (\dA+\bw,13.77);
\fill[barEng] (\dB-\bw,0) rectangle (\dB+\bw,7.38);
\fill[barOwn] (\dC-\bw,0) rectangle (\dC+\bw,15.75);
\fill[barEng] (\dD-\bw,0) rectangle (\dD+\bw,12.69);
\fill[barEng] (\dE-\bw,0) rectangle (\dE+\bw,10.44);
\fill[barEng] (\dF-\bw,0) rectangle (\dF+\bw,13.95);
\foreach \cx/\h/\v in {\dA/13.77/15.3, \dB/7.38/8.2, \dC/15.75/17.5, \dD/12.69/14.1, \dE/10.44/11.6, \dF/13.95/15.5}
   { \node[tick, anchor=south] at (\cx, \h+0.5) {\v}; }
\draw[ax] (69,0) -- (121,0);
\draw[ax] (69,0) -- (69,19.5);
\foreach \v/\yy in {0/0, 10/9, 20/18} { \node[tick, anchor=east] at (68, \yy) {\v}; }
\node[vlab, rotate=90, anchor=south, align=center] at (62, 9) {per-decision lift\\(\% over no branching)};
\node[font=\sffamily\tiny, text=pxOwn, anchor=south east] at (120.5, 15.95) {stochastic (policy's own)};
\node[font=\sffamily\tiny\itshape, text=pxSlate, anchor=south] at (95, 18.3) {lift does not follow};
\foreach \cx/\name in {%
    \dA/{Tempered $T{=}2$}, \dB/{LLM-prefix $\ell{=}3$},
    \dC/{Stochastic $\star$}, \dD/{Top-$K$ distinct},
    \dE/{LLM-prefix $\ell{=}1$}, \dF/{Counterfactual}}
  { \node[font=\sffamily\tiny, text=pxInk, rotate=32, anchor=north east] at (\cx+1.3, -1) {\name}; }

\end{tikzpicture}
  \caption{\textbf{Engineered dispersion does not increase lift.} Six proposer conditions against the value-head verifier, ordered by the spread they produce. \textbf{Left:} proposal engineering moves $\Delta_c$ by roughly a factor of two ($0.38 \to 0.86$). \textbf{Right:} lift does not follow, and no engineered proposer beats the policy's own stochastic one (teal, $\star$). Per-condition numbers in Table~\ref{tab:b2_proposer}.}
  \Description{Two side-by-side panels over six proposer conditions ordered by dispersion. The left panel shows dispersion roughly doubling across engineered proposers. The right panel shows that lift does not rise with dispersion, and the policy's own stochastic proposer, marked with a star, has the highest lift.}
  \label{fig:proposer_axis}
\end{figure*}

Companion to Section~\ref{subsec:b2} and Figure~\ref{fig:proposer_axis}. Table~\ref{tab:b2_proposer} reports per-condition lift, directly-measured per-decision dispersion, and independently-measured verifier rank fidelity for the five proposer mechanisms. Dispersion moves by a factor of two across mechanisms while rank fidelity stays at $0.60$ to $0.67$, so the verifier exploits the engineered spread, and episode return still does not rise above the stochastic reference. Individual deficits depend on which stochastic pool serves as the reference, since the same configuration reads $37.04$ in the ten-episode proposer sweep, $36.00 \pm 0.56$ in the clean rerun of Appendix~\ref{app:pools}, and $36.21$ in the fifty-episode pool of Section~\ref{sec:investigation}, and none is significant at three seeds. The claim therefore rests on all five mechanisms falling below the stochastic reference (sign test, one-sided $p = 0.031$), which holds under every choice of pool.

\begin{table*}[t]
  \centering
  \caption{Cross-mechanism proposal ablation. $K{=}7$, $H{=}20$, periodic-50, $\pi_{\text{strong}}$ on Craftax-Symbolic-Medium with the value-head verifier. Three seeds per cell except $\dagger$ ($2$ seeds), at $10$ episodes per seed, matched to the stochastic proposer which was only run at that budget. $\Delta_c$ is measured per-decision dispersion (parenthetical: shift from stochastic); rank fidelity $\bar\rho$ is the within-decision Spearman correlation between the value-head score and realized $H$-step return, distinct from the top-1 accuracy $\alpha$ of the main text. Bold marks the stochastic baseline, which no mechanism beats.}
  \label{tab:b2_proposer}
  \footnotesize
  \begin{tabular}{lccccc}
    \toprule
    Proposal mechanism & Episode return & Lift & $\Delta_c$ & Rank fidelity $\bar\rho$ & Return $-$ stoch \\
    \midrule
    No branching                            & $32.09$ & --- & --- & --- & --- \\
    Stochastic ($K \sim \pi$)               & $\mathbf{37.04 \pm 0.20}$ & $\mathbf{+15.4\%}$ & $0.51$ (ref) & $0.599$ & --- \\
    \midrule
    B1 Top-$K$ distinct                     & $35.97 \pm 2.61$ & $+12.1\%$ & $0.52$ ($+1\%$)   & $0.621$ & $-1.07$ \\
    B1 Tempered ($T{=}2.0$)                 & $36.34 \pm 1.18$ & $+13.2\%$ & $0.38$ ($-25\%$)  & $0.667$ & $-0.70$ \\
    B1 Counterfactual$^{\dagger}$           & $36.41 \pm 0.40$ & $+13.5\%$ & $0.86$ ($+69\%$)  & $0.622$ & $-0.63$ \\
    \midrule
    B2 LLM-prefix $L{=}1$                & $35.20 \pm 3.67$ & $+9.7\%$ & $0.65$ ($+28\%$)  & $0.623$ & $-1.84$ \\
    B2 LLM-prefix $L{=}3$                & $34.13 \pm 0.15$ & $+6.4\%$  & $0.44$ ($-15\%$)  & $0.653$ & $-2.91$ \\
    \bottomrule
  \end{tabular}
\end{table*}

\subsection{Gating Procedure and Selection-Set Evaluation}
\label{app:dispersion_targeted_extra}

Algorithm~\ref{alg:method} states the deployment-time procedure, with the gating signal $g$ the only swapped component. We chose the threshold on three selection seeds (Table~\ref{tab:dispersion_targeted_online}) and then evaluated it on twelve fresh seeds (Table~\ref{tab:gating_heldout}). The held-out evaluation is the primary evidence; the selection-set results below show the curve the threshold was chosen from.

\begin{algorithm}[t]
\caption{Dispersion-Gated Inference-Time Branching.
  The gating signal $g$ admits three instantiations: periodic-$T$ ($g(s){=}\mathbf{1}[t\bmod T{=}0]$, no training needed), value magnitude ($g(s){=}|\hat V(s)|$), or the learned predictor ($g{=}\hat{\Delta}_\theta$, requires the offline fit in lines~\ref{line:collect}--\ref{line:calibrate}).}
\label{alg:method}
\begin{algorithmic}[1]
\Statex \textbf{Input:} policy $\pi$; value function $V$; gating signal $g$; threshold $\tau$; rollout count $K$; horizon $H$
\If{$g = \hat{\Delta}_\theta$} \hfill\Comment{offline, once; skip for periodic-$T$ or $|V(\cdot)|$}
    \State Collect $\{(s_i,\Delta(s_i))\}$ by running $\pi$ with periodic-$T$ branching \label{line:collect}
    \State Fit $\hat{\Delta}_\theta$ via ridge regression (5-fold CV); calibrate $\tau$ to target trigger rate $b$ \label{line:calibrate}
\EndIf
\For{each episode}
    \State $s \gets s_0$
    \For{each step $t$}
        \If{$g(s) > \tau$}
            \State Sample $a_k \sim \pi(\cdot \mid s)$ for $k = 1,\ldots,K$
            \State Rollout each $a_k$ for $H$ steps; obtain trajectory $h_k$
            \State $k^* \gets \arg\max_k\bigl[R(h_k) + \gamma^H \hat V(s_H^{(k)})\bigr]$
            \State $s \gets s_H^{(k^*)}$
        \Else
            \State $s \gets \mathrm{Step}(s,\;a \sim \pi(\cdot \mid s))$
        \EndIf
    \EndFor
\EndFor
\end{algorithmic}
\end{algorithm}

\begin{table}[t]
  \centering
  \caption{Selection-set evaluation of dispersion-gated triggering against uniform periodic triggering on Craftax-Medium, for two policy architectures, three seeds $\times$ ten episodes per row. Lift is over the per-policy no-branching baseline, and Lift/call divides it by calls per episode. Bold marks each policy's most call-efficient gated row. The GTrXL periodic-$50$ row is the historical aggregate (Appendix~\ref{app:pools}).}
  \label{tab:dispersion_targeted_online}
  \footnotesize
  \resizebox{\columnwidth}{!}{\begin{tabular}{llcccc}
    \toprule
    Policy & Method & Calls/ep & Episode return & Lift & Lift/call \\
    \midrule
    \multirow{4}{*}{GTrXL ($\pi_{\text{strong}}$)}
      & No branching                          & $0$    & $32.09$ & --- & --- \\
      & Periodic-50 (uniform)                 & $7.0$  & $35.85 \pm 1.88$ & $+11.7\%$ & $1.67\%$ \\
      & Disp-target $\tau{=}3.0$               & $3.1$  & $34.50 \pm 1.55$ & $+7.5\%$ & $\mathbf{2.42\%}$ \\
      & Disp-target $\tau{=}1.5$               & $9.6$  & $36.00 \pm 3.86$ & $+12.2\%$ & $1.27\%$ \\
    \midrule
    \multirow{4}{*}{PPO-MLP}
      & No branching                          & $0$    & $5.37 \pm 0.17$ & --- & --- \\
      & Periodic-50 (uniform)                 & $10.4$ & $7.23 \pm 0.18$ & $+34.6\%$ & $3.33\%$ \\
      & Disp-target $\tau{=}0.73$              & $9.4$ & $7.30 \pm 0.21$ & $+35.9\%$ & $\mathbf{3.82\%}$ \\
      & Disp-target $\tau{=}0.20$              & $45.9$ & $8.77 \pm 0.09$ & $+63.3\%$ & $1.38\%$ \\
    \bottomrule
  \end{tabular}}
\end{table}

\paragraph{Dense $\tau$-sweep: per-call efficiency curve.}
The two predictor-gated rows of Table~\ref{tab:dispersion_targeted_online} (GTrXL block) are anchors on a denser threshold sweep, summarized in Table~\ref{tab:tau_sweep}. Three observations: \emph{(i)} per-call efficiency (lift per branching call, last column) is highest at the most aggressive gating and falls as the call budget grows, non-monotonically at adjacent thresholds ($\tau{=}2.0$'s $0.99\%$/call sits below $\tau{=}1.5$'s $1.27\%$/call, an inversion within the seed noise of observation \emph{(ii)}), the qualitative saturation the dispersion bound predicts when the predictor is concentrating budget on the high-dispersion tail. \emph{(ii)} Absolute lift is non-monotone in calls/episode at the seed-noise level (SEM on episode return is $\sim\!2$ return units per row, i.e.\ $\sim\!6$\,pp on the lift scale, so the $\tau{=}2.0$ to $\tau{=}3.0$ ordering and the $\tau{=}1.5$ vs.\ periodic-50 ordering are both inside the noise). The headline operating point $\tau{=}3.0$ is selected as the highest-lift row that strictly dominates periodic-50 on per-call efficiency. \emph{(iii)} The single very-aggressive point $\tau{=}1.0$ ($14.0$ calls/episode, return $30.33$, lift $-5.5\%$) is a stress test of the over-branching regime, called out separately in the next paragraph rather than mixed into the efficiency curve.

\begin{table}[!htbp]
  \centering
  \caption{Online dispersion-gated triggering on Craftax-Medium $\pi_{\text{strong}}$ (GTrXL), full $\tau$-sweep. Three seeds $\times$ 10 episodes per row. Baseline is the Craftax-Medium no-branching pool ($32.09$; Section~\ref{sec:methodology}). $\tau{=}1.5$ and $\tau{=}3.0$ are the rows surfaced in Table~\ref{tab:dispersion_targeted_online}; $\tau{=}2.0$ and $\tau{=}4.0$ fill in the curve between them. Per-call efficiency is lift (\%) divided by calls/episode; periodic-50 is the budget-matched uniform reference. Predictor and threshold semantics are identical across all four $\tau$ rows (same $\hat\Delta_\theta$).}
  \label{tab:tau_sweep}
  \footnotesize
  \resizebox{\columnwidth}{!}{\begin{tabular}{lccccc}
    \toprule
    Method & Calls/ep & Episode return & Lift & Per-call efficiency \\
    \midrule
    No branching                    & $0$    & $32.09$ & ---       & --- \\
    Disp-target $\tau{=}4.0$        & $1.23$ & $33.40$          & $+4.1\%$  & $3.32\%$/call \\
    Disp-target $\tau{=}3.0$ (headline) & $3.1$  & $34.50 \pm 1.55$ & $\mathbf{+7.5\%}$ & $\mathbf{2.42\%}$/call \\
    Disp-target $\tau{=}2.0$        & $6.23$ & $34.07$          & $+6.2\%$  & $0.99\%$/call \\
    Periodic-50 (uniform)           & $7.0$  & $35.85 \pm 1.88$ & $+11.7\%$ & $1.67\%$/call \\
    Disp-target $\tau{=}1.5$        & $9.6$  & $36.00 \pm 3.86$ & $+12.2\%$ & $1.27\%$/call \\
    \bottomrule
  \end{tabular}}
\end{table}

\begin{figure}[t]
  \centering
  \includegraphics[width=0.6\linewidth]{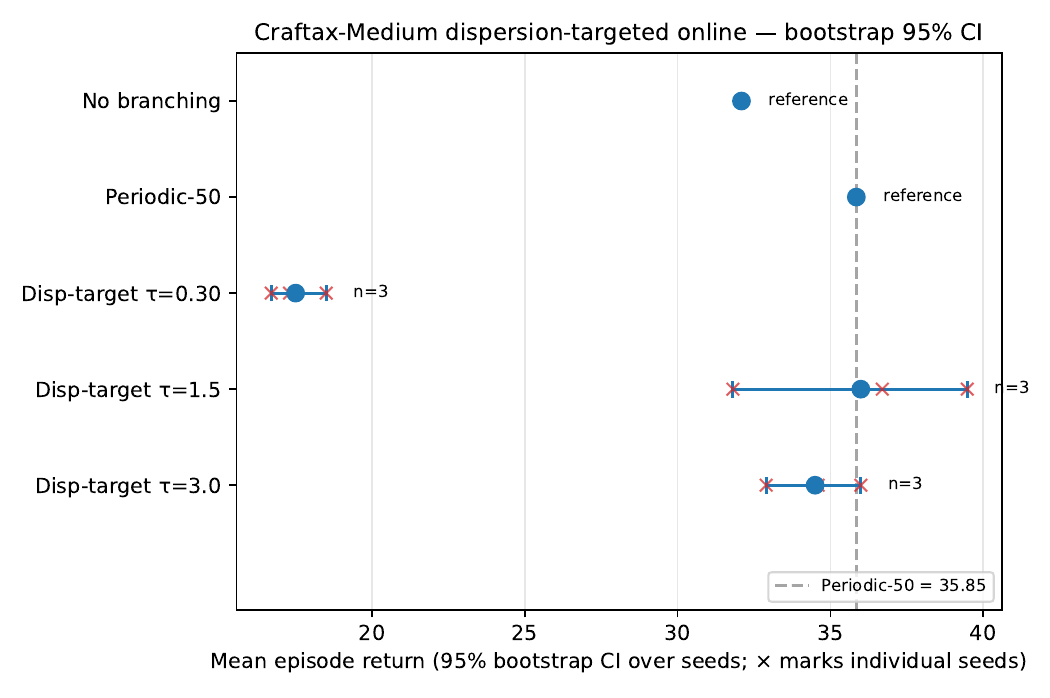}
  \caption{\textbf{Bootstrap 95\% confidence intervals over seeds for Table~\ref{tab:dispersion_targeted_online} on Craftax-Medium.} Per-condition seed-level mean returns are aggregated by $10{,}000$-iteration bootstrap; $\times$ marks individual seeds. On the selection seeds, $\tau{=}3.0$ retains roughly two thirds of periodic-50's lift at $44\%$ of its calls, with overlapping intervals. The $\tau{=}0.30$ row is a further over-branching threshold, below the $\tau{=}1.0$ stress test of this appendix, whose return collapses to $17.5$. The no-branching marker is the canonical $32.09$ pool of Appendix~\ref{app:pools}. DoorKey-8x8 is omitted because its policy is at the reward ceiling (Table~\ref{tab:cross_env_dispersion}).}
  \Description{Interval plot of bootstrap 95 percent confidence intervals of mean episode return on Craftax-Medium for each branching condition, with crosses marking individual seeds and a reference marker for the no-branching baseline. The gated condition at threshold 3.0 has an interval overlapping periodic-50.}
  \label{fig:bootstrap_table5}
\end{figure}

\paragraph{Cluster bootstrap on the per-call-efficiency ratio.} This paragraph bounds the resolution of the \emph{selection-set} estimate; the held-out evaluation on seeds $45$--$56$ in Section~\ref{sec:dispersion_gating} is the primary evidence and gives $1.45\times$ with a seed-level $95\%$ interval of $[0.90, 2.28]$ and $P(\text{ratio}>1) = 0.94$ over the twelve matched seeds. The two arms there return $35.17 \pm 0.73$ at $3.34$ measured calls per episode under gating, against $36.54 \pm 0.77$ at the protocol's nominal $7.0$ under uniform branching. The selection-set figure of Table~\ref{tab:dispersion_targeted_online} happens to land at the same $1.45\times$, but it is a ratio of two lift-per-call quantities each from three seeds, so we bound its resolution separately here. Per-episode returns and branching-call counts are resampled jointly within each seed, independently per arm, over $10{,}000$ iterations, with calls per episode re-estimated each iteration and the baseline cancelling. Across specifications (varying the periodic-$50$ reference and using either the canonical baseline $32.09$ or a seed-matched $33.60$) the point estimate runs from $0.99$ to $2.04$, every $95\%$ interval spans parity, and $P(\text{ratio}>1)$ runs from $0.51$ to $0.77$; the interval is wide because the denominator can approach zero under resampling. Two caveats belong with it. The periodic-$50$ row aggregates a historical pool mixing selector variants and episode counts; rerunning that arm at the gated protocol (value head, $K{=}7$, $H{=}20$, periodic-$50$, three seeds $\times$ ten episodes) returns $36.70$, $34.90$, $36.40$, mean $36.00 \pm 0.56$, so the historical value is not a pool artifact, though clean matched-seed reconstructions fire at $7.5$--$9.5$ calls per episode rather than $7.0$. We therefore treat the multiplier as unresolved and rest Section~\ref{sec:dispersion_gating}'s claim on the retained-lift-at-reduced-calls comparison (Figure~\ref{fig:bootstrap_table5}) and the Asterix scope prediction.

\paragraph{Over-branching stress test ($\tau{=}1.0$).}
At $\tau{=}1.0$ the predictor fires at $14.0$ calls/episode (twice the periodic-50 budget); mean episode return is $30.33$ (3 seeds: $30.5$, $29.2$, $31.3$), lift $-5.5\%$ relative to the no-branching baseline. The drop below baseline at high call count is the off-policy first-action tax (Section~\ref{subsec:b2}, Appendix~\ref{app:slope_interpretation}) made directly visible: every additional triggered branch commits the next step to one of $K$ stochastically-sampled (i.e.\ off-policy from $\pi$'s argmax) candidates, and once branching is firing on the bulk of the dispersion distribution (not just the tail) the accumulated tax outweighs the marginal selection gain. Mechanistically this is the same downstream tax that nullifies the proposer axis on the candidate-side (Appendix~\ref{app:b2_proposer}); here it sets an upper bound on the budget side of the dispersion-gating recipe. The over-branching regime is a stress test, not part of the deployment curve.

\subsection{Trigger Baselines: Anti-Targeting and Sparse Periodic}

Budget-matched comparison of triggering strategies under the value-head verifier, isolating \emph{which} decisions to branch on from the choice of verifier. The table supplies the value-only \emph{anti-targeting} row ($-8.6\%$ lift) and the periodic-200 row anchoring the per-call comparison in Table~\ref{tab:dispersion_targeted_online}, which fails to clear the corrected no-branching baseline.

\begin{table}[t]
  \centering
  \caption{Budget-matched triggering comparison. $K{=}7$, $H{=}20$, $\pi_{\text{strong}}$ (GTrXL) on Craftax-Symbolic-Medium with the value head as the verifier across all rows. Calls/episode is the empirical mean across 3 seeds × 10 episodes; periodic-50 is the historical aggregate. Baseline is the Craftax-Medium no-branching pool ($32.09$; Section~\ref{sec:methodology}). Bold marks the anti-targeting row. The $\hat V<5.0$ rule fires whenever the value head drops below the empirical threshold $5.0$ on Craftax-Medium (the value head is dimensionless on this env), chosen to fire at a rate comparable to periodic-50.}
  \label{tab:triggering}
  \footnotesize
  \resizebox{\columnwidth}{!}{\begin{tabular}{lccc}
    \toprule
    Trigger Strategy & Episode Return & Calls/Ep. & Lift vs.\ baseline \\
    \midrule
    No branching                                  & $32.09$ & $0$    & ---     \\
    Periodic (period $50$, historical)             & $35.85 \pm 1.88$ & $\sim 7$ & $+11.7\%$ \\
    Periodic (period $200$)                        & $31.73 \pm 1.25$ & $1.8$  & $-1.1\%$ \\
    \midrule
    \multicolumn{4}{l}{\emph{Anti-targeting baseline:}} \\
    Value only ($\hat V<5.0$)                           & $\mathbf{29.34 \pm 0.45}$ & $\mathbf{8.2}$ & $\mathbf{-8.6\%}$ \\
    \bottomrule
  \end{tabular}}
\end{table}

\paragraph{Anti-targeting hurts: value-only triggering drops below baseline.} Value-only ($\hat V<5.0$) fires often ($8.2$ calls/ep) yet achieves $29.34$, $-8.6\%$ below the no-branching baseline. Mechanism: $\hat V<5.0$ holds at low-stakes early-game states where $\Delta(s)$ is also small, so value-only routes oracle queries to exactly the decisions where they matter least. \emph{Anti-targeting on an anti-dispersion signal can underperform no-branching at all.} The control is a structural sanity check: the gating signal must be dispersion-correlated to be safe; cf.\ \S\ref{app:pareto_baselines} where $|\hat V(s)|$ as a magnitude proxy is $+\rho$ with $\Delta$, while the value-threshold gate here is anti-correlated within the low-value regime.

\paragraph{Periodic-200 anchor for the per-call-efficiency comparison.} Periodic-200 at $\sim 1.8$ calls/ep gets $-1.1\%$ lift, failing to clear the corrected no-branching baseline; the dispersion-targeted recipe at $\tau{=}3.0$ in Table~\ref{tab:dispersion_targeted_online} achieves $+7.5\%$ lift at $3.1$ calls/ep, so a per-call-matched periodic-200 sweep does not provide a positive baseline to gate against, as discussed in \S\ref{subsec:dispersion_targeted}.

\subsection{Hindsight Pareto Across Cheap Pre-Branch Signals}
\label{app:pareto_baselines}

Companion to Section~\ref{sec:dispersion_gating}. On the pooled $n{=}1{,}780$ decisions used to fit the predictor, its top-$30\%$ predicted decisions capture $68.0\%$ of $\sum_s \Delta(s)$, against $30.4\%$ for random gating ($2.24\times$). Table~\ref{tab:pareto_baselines} breaks hindsight capture down by budget and gating signal on the paper-protocol dumps of each environment ($n{=}977$ on Craftax-Medium), and four patterns stand out. First, policy entropy is uncorrelated with $\Delta$ in every environment ($\rho \in [-0.07, +0.05]$). Second, on Craftax-Medium the learned ridge predictor matches or beats $|\hat V(s)|$ at every budget, by $12$ points at $B{=}33\%$. Third, on Asterix the predictor and $|\hat V(s)|$ exchange the lead, with $|\hat V(s)|$ ahead at mid-to-high budgets, consistent with both being noisy proxies for $\sigma(s)$. Fourth, on the zero-dispersion R9 and DoorKey environments no gate captures anything, since there is nothing to capture.

\begin{table*}[t]
  \centering
  \caption{Hindsight Pareto: fraction of total $\Delta$-mass captured at
    compute budget $B$ across the four environments along the regime axis.
    Each block header gives the Spearman $\rho$ of each signal against realized $\Delta$. Bold marks the best
    non-oracle gate per row. The Asterix concentration is computed on the original Asterix decision
    dump ($n{=}454$); the reward-masking fix of Appendix~\ref{app:methodological_audit}
    shifted Asterix $\Delta_c$ by less than sampling noise ($0.521 \to 0.527$),
    so the concentration ratios are unaffected.}
  \label{tab:pareto_baselines}
  \footnotesize
  \begin{tabular}{lrrrrrr}
    \toprule
    Budget $B$ & $n$ & Random & Entropy & $|\hat V(s)|$ & Predicted & Oracle \\
    \midrule
    \multicolumn{7}{l}{\emph{Craftax-Symbolic-Medium ($\pi_{\text{strong}}$, GTrXL):} $\rho_{\text{ent}}{=}+0.05$, $\rho_{|\hat V|}{=}+0.21$, $\rho_{\text{pred}}{=}+0.41$} \\
    10\% & 977 & $10\%$ & $13\%$ & $39\%$ & $\mathbf{39\%}$ & $63\%$ \\
    33\% & 977 & $33\%$ & $31\%$ & $58\%$ & $\mathbf{70\%}$ & $97\%$ \\
    50\% & 977 & $50\%$ & $46\%$ & $70\%$ & $\mathbf{86\%}$ & $100\%$ \\
    \midrule
    \multicolumn{7}{l}{\emph{Asterix-MinAtar ($\pi_{\text{MLP}}$):} $\rho_{\text{ent}}{=}+0.02$, $\rho_{|\hat V|}{=}+0.21$, $\rho_{\text{pred}}{=}+0.14$} \\
    10\% & 454 & $10\%$ & $12\%$ & $12\%$ & $\mathbf{12\%}$ & $26\%$ \\
    33\% & 454 & $33\%$ & $33\%$ & $38\%$ & $\mathbf{39\%}$ & $63\%$ \\
    50\% & 454 & $50\%$ & $50\%$ & $\mathbf{56\%}$ & $55\%$ & $87\%$ \\
    70\% & 454 & $70\%$ & $71\%$ & $\mathbf{80\%}$ & $72\%$ & $100\%$ \\
    \midrule
    \multicolumn{7}{l}{\emph{R9-19x19 and DoorKey-8x8 ($\pi_{\text{MLP}}$): zero-dispersion regime post-bugfix ($\Delta_c{=}0.00$ and $\approx 0.03$); no tail to gate.}}\\
    any & 1{,}990 & $0\%$ & $0\%$ & $0\%$ & $0\%$ & $0\%$ \\
    \bottomrule
  \end{tabular}
\end{table*}

\subsection{Permuted-Gate Negative Control}
\label{app:gate_permutation}

Hindsight test that the gate's per-call efficiency comes from per-state alignment with $\Delta(s)$ and not from a marginal-distribution artifact. On the paper-protocol Craftax-Medium GTrXL dumps (periodic-50, $K{=}7$, $H{=}20$; $977$ decisions over $120$ episodes, pool mean $\Delta = 0.527$) we shuffle the gating signal across decisions \emph{within each episode}, then rank and capture the top-$B$. That preserves the signal's marginal and the firing rate $B$ while destroying per-state alignment, so the gating claim predicts collapse to the random baseline. It collapses. At $B{=}3$, matching the headline firing rate, $|\hat V(s)|$ captures $0.950$ against random $0.811$ ($1.17\times$) and falls to $0.812$ when permuted; the learned $\hat\Delta_\theta$ shows the same pattern at higher amplitude, $1.190$ ($1.47\times$) against $0.806$ permuted ($0.99\times$). Script: \texttt{scripts/analysis/gate\_permutation\_control.py}.

\begin{table}[!htbp]
  \centering
  \caption{Permuted-gate hindsight control: mean captured $\Delta$ at top-$B$ decisions per episode. \textbf{Actual}: rank within each episode by the gating signal, capture top-$B$. \textbf{Permuted}: shuffle the signal within each episode first. \textbf{Random}: pick $B$ decisions uniformly. Permuted and random are mean $\pm$ SEM over $200$ resamples. At the headline $B{=}3$, permutation collapses both signals to random, so the gate's efficiency is carried by per-state alignment and not by the signal's marginal.}
  \label{tab:gate_permutation}
  \footnotesize
  \resizebox{\columnwidth}{!}{\begin{tabular}{rcccc}
    \toprule
    $B$ & Actual gate & Permuted (within-ep) & Random & Concentration (actual/random) \\
    \midrule
    \multicolumn{5}{l}{\textit{$|\hat V(s)|$ as gating signal:}} \\
    $1$ & $1.708$ & $0.868\,{\pm}\,0.008$ & $0.849\,{\pm}\,0.008$ & $2.01\times$ \\
    $2$ & $1.142$ & $0.855\,{\pm}\,0.005$ & $0.847\,{\pm}\,0.004$ & $1.35\times$ \\
    $\mathbf{3}$ & $\mathbf{0.950}$ & $\mathbf{0.812\,{\pm}\,0.003}$ & $\mathbf{0.811\,{\pm}\,0.003}$ & $\mathbf{1.17\times}$ \\
    $4$ & $0.867$ & $0.764\,{\pm}\,0.002$ & $0.759\,{\pm}\,0.002$ & $1.14\times$ \\
    $5$ & $0.791$ & $0.725\,{\pm}\,0.001$ & $0.725\,{\pm}\,0.002$ & $1.09\times$ \\
    $7$ & $0.718$ & $0.679\,{\pm}\,0.001$ & $0.678\,{\pm}\,0.001$ & $1.06\times$ \\
    \midrule
    \multicolumn{5}{l}{\textit{$\hat\Delta_\theta(s)$ (learned ridge) as gating signal:}} \\
    $1$ & $1.946$ & $0.873\,{\pm}\,0.008$ & $0.849\,{\pm}\,0.008$ & $2.29\times$ \\
    $2$ & $1.371$ & $0.855\,{\pm}\,0.004$ & $0.847\,{\pm}\,0.004$ & $1.62\times$ \\
    $\mathbf{3}$ & $\mathbf{1.190}$ & $\mathbf{0.806\,{\pm}\,0.002}$ & $\mathbf{0.811\,{\pm}\,0.003}$ & $\mathbf{1.47\times}$ \\
    $4$ & $1.007$ & $0.761\,{\pm}\,0.002$ & $0.759\,{\pm}\,0.002$ & $1.33\times$ \\
    $5$ & $0.881$ & $0.725\,{\pm}\,0.001$ & $0.725\,{\pm}\,0.002$ & $1.21\times$ \\
    $7$ & $0.741$ & $0.679\,{\pm}\,0.001$ & $0.678\,{\pm}\,0.001$ & $1.09\times$ \\
    \bottomrule
  \end{tabular}}
\end{table}

\subsection{Recipe Transfer Across Environments and Policies}
\label{app:cross_env_transfer}

Companion to the cross-environment discussion in Section~\ref{sec:dispersion_gating}. The recipe is refit independently on each environment's periodic-branching dumps using the same ridge regression on the standardized raw observation; the cheap $|\hat V(s)|$ proxy is the absolute mean of the bootstrapped branch scores from the same dumps. Both are positive on every environment with non-zero realized dispersion, strongest on rich-state Craftax-Medium and weakest on Craftax-Hard, where dispersion is near the floor and the form predicts little upside even from a perfect gate. Ridge-on-observation therefore extracts a positive signal everywhere we test it, while the per-call efficiency \emph{magnitude} depends on whether the environment's dispersion distribution has a tail to gate against. The proxy reads higher than the learned recipe in all three rows of Table~\ref{tab:cross_env_transfer}, which are the two $\pi_{\text{weak}}$ cells and Craftax-Hard. On the GTrXL Craftax-Medium cell that carries the gating result, the order reverses and the learned predictor leads (Table~\ref{tab:pareto_baselines}). We therefore scope the predictor to states rich enough to support it and take $|\hat V(s)|$ as the default elsewhere.

\begin{table}[!htbp]
  \centering
  \caption{Cross-environment audit of the dispersion-gating recipe. Decision dumps are the same paper-protocol periodic-branching runs underlying the scaling-law fit. Recipe Spearman $\rho$ is 5-fold CV on the ridge-on-observation predictor refit per environment; $|\hat V(s)|$-proxy $\rho$ is the cross-decision Spearman between $|\text{mean}_k\,s_k|$ (where $s_k$ are the bootstrapped branch scores) and realized $\Delta(s)$.}
  \label{tab:cross_env_transfer}
  \footnotesize
  \begin{tabular}{lrrr}
    \toprule
    Environment & $n_{\text{decisions}}$ & Recipe CV $\rho$ & $|V|$-proxy $\rho$ \\
    \midrule
    Craftax-Medium (MLP)  & $313$  & $+0.19$ & $+0.44$ \\
    Craftax-Hard          & $831$  & $+0.09$ & $+0.18$ \\
    Asterix-MinAtar       & $1{,}468$ & $+0.20$ & $+0.31$ \\
    \bottomrule
  \end{tabular}
\end{table}

Table~\ref{tab:asterix_gating} is the online counterpart: on Asterix-MinAtar, where the top $30\%$ of decisions hold only $1.14\times$ the random share of dispersion mass, every gated configuration underperforms uniform periodic-15 at matched or lower cost, which is the form's scope prediction.

\begin{table}[!htbp]
  \centering
  \caption{Asterix-MinAtar online gating. Three seeds $\times$ 30 episodes per branched row; the no-branching row is the paired baseline of Table~\ref{tab:cross_env_dispersion}. Lift is over the no-branching baseline; Lift/call is lift divided by per-episode call count. The two highest-firing gated configurations match periodic-$15$'s Lift/call ($0.87$ and $0.89$ against $0.89$) while the two sparsest fall below it, as expected when there is no tail to gate onto. Contrast Table~\ref{tab:dispersion_targeted_online} on Craftax-Medium.}
  \label{tab:asterix_gating}
  \footnotesize
  \begin{tabular}{lcccc}
    \toprule
    Method & Calls/ep & Episode return & Lift & Lift/call \\
    \midrule
    No branching (paired eval)       & $0$    & $1.37 \pm 0.08$  & ---       & --- \\
    Periodic-15 (uniform)            & $16.3$ & $15.82 \pm 0.41$ & $+14.45$  & $0.89$ \\
    \midrule
    $|V|$-gated $\tau{=}0.92$        & $5.0$  & $5.70 \pm 0.50$  & $+4.33$   & $0.87$ \\
    $|V|$-gated $\tau{=}1.30$        & $1.4$  & $2.31 \pm 0.27$  & $+0.94$   & $0.67$ \\
    $|V|$-gated $\tau{=}1.66$        & $0.2$  & $1.17 \pm 0.13$  & $-0.20$   & $-1.00$ \\
    Predictor-gated $\tau{=}0.6$     & $4.2$  & $5.10 \pm 0.83$  & $+3.73$   & $0.89$ \\
    \bottomrule
  \end{tabular}
\end{table}

\paragraph{Cross-policy replication (PPO-MLP).} A separately-fit MLP dispersion predictor (same recipe; train Spearman $\rho{=}+0.73$, top-third capture $92.7\%$, $n_{\text{train}}{=}805$) drives predictor-gated branching at similar compute ($\tau{=}0.73$, $9.4$ calls/ep) to $+35.9\%$ lift against periodic-50's $+34.6\%$ at $10.4$ calls/ep, a per-call efficiency of $1.15\times$ (Table~\ref{tab:dispersion_targeted_online}). The smaller magnitude than GTrXL's $1.45\times$ is consistent with the weaker policy's episodes being denser in low-stakes decisions, compressing the tail. The direction replicates across architectures; the ratio does not resolve.

\clearpage

\else

\section{Introduction}
\label{sec:introduction}

A common constraint in deploying reinforcement learning agents is that the
policy cannot be changed. Retraining may require data or computational
resources that are no longer available, and a deployed policy may be
externally supplied, validated, or too costly to update. An operator may
nevertheless control how the policy is executed. In digital environments
whose states can be saved and restored, the operator can run the policy
forward several times from the same state, evaluate the resulting
continuations, and resume from the endpoint of the most promising one.
This offers a way to improve the execution of a frozen policy using
additional computation at deployment time.

We call this procedure \emph{branch-and-commit}. It applies when the
environment state can be copied and resumed, as in emulated games and
sandboxed software systems; we do not consider physical-world deployment.
Within this setting, we ask three questions: when does branch-and-commit
improve return, what produces the improvement, and where should its
computational budget be spent?

Branching can help only when the sampled continuations differ in value. If
all candidates produce similar returns, selecting among them offers little
benefit, regardless of how accurately they are evaluated. We measure this
opportunity using \emph{dispersion}, defined as the difference between the
best candidate return and the mean candidate return. High dispersion
indicates that branching has produced a potentially useful continuation;
low dispersion leaves little room for selection to improve on an ordinary
policy rollout.

The source of this dispersion also matters. Candidate branches may begin
with different actions, allowing the operator to select a better action.
Alternatively, they may begin with the same action but encounter different
stochastic transitions. Branching then selects among different realizations
of a decision the policy has already made. We call these mechanisms
\emph{action selection} and \emph{realization selection}, respectively.
The distinction changes how an improvement should be interpreted: higher
return does not necessarily mean that additional computation caused the
agent to choose better actions.

Branch-and-commit is related to test-time compute methods that generate and
evaluate multiple candidate solutions. In language models, the success of
these methods often depends on the verifier used to select among
candidates~\cite{brown2024large,lightman2023lets}. A simulator, however,
already reveals the rewards received during each continuation. In this
setting, the amount and source of variation among candidates, and the
decision of when to generate them, may matter as much as the sophistication
of the verifier.

We study branch-and-commit across four reinforcement learning environments,
two frozen policies, five continuation verifiers, a model-free selection
rule, and a range of branching
configurations. Dispersion predicts when branching improves episode return.
Branching provides no benefit in low-dispersion settings, including two
identified in advance, while increasing return by $12.8\%$ at the
main operating point in Craftax-Medium. Most of this benefit comes from
realization selection. Under the stronger Craftax policy, all candidates
choose the same first action at $84.9\%$ of branching decisions, and an
intervention that shares environment randomness across candidates
substantially reduces dispersion.

Verifier choice has a smaller effect at this operating budget. A value
head, GPT-4o, and a model-free rule that selects the largest realized
continuation return achieve similar episode returns. In contrast, deciding
where to branch substantially affects computational cost. We train a
predictor of dispersion from the current observation and use it to gate
branching before any candidates are generated. On twelve held-out seeds,
dispersion gating retains $69\%$ of the improvement from uniform periodic
branching while using $48\%$ as many branching calls.

Our contributions are fourfold. First, we formalize branch-and-commit for
frozen reinforcement learning policies and use dispersion to characterize
when branching has the potential to help. Second, we distinguish action
selection from realization selection and show that realization selection
accounts for most of the gain in our main setting. Third, we show that
predicted dispersion can guide the allocation of a limited branching
budget. Fourth, we show that test-time branching can raise a checkpoint's
reported return without changing its policy, so evaluations should report
whether gains come from action or realization selection. Together, the results explain when branch-and-commit is useful,
what its gains represent, and where its computation should be spent.

\section{Branch-and-Commit and Candidate Dispersion}
\label{sec:problem}

We consider a fixed Markov decision process
$\mathcal{M}=(\mathcal{S},\mathcal{A},P,r,\gamma)$ and a frozen policy $\pi$.
The operator cannot retrain $\pi$, but can fork the environment from its
current state and resume execution from a selected simulated continuation.
Transitions may be stochastic because the selected continuation is retained
rather than replayed. Figure~\ref{fig:branch_and_commit_overview} summarizes the procedure and
the two possible sources of improvement.

\begin{figure}[t]
    \centering
    \resizebox{\columnwidth}{!}{\definecolor{bcS1}{HTML}{5B6470}\definecolor{bcS2}{HTML}{C98A1B}
\definecolor{bcS3}{HTML}{2E8B7A}\definecolor{bcS4}{HTML}{B23A48}
\definecolor{bcAct}{HTML}{2F6DB5}\definecolor{bcReal}{HTML}{7A4FB0}
\begin{tikzpicture}[
    font=\sffamily\scriptsize,
    >={Stealth[length=1.5mm,width=1.2mm]},
    card/.style={draw=#1!70, fill=white, rounded corners=1pt, line width=0.5pt,
        align=center, inner sep=2.2pt, minimum width=1.6cm, minimum height=0.74cm},
    card/.default=black,
    tab/.style={fill=#1, text=white, font=\sffamily\scriptsize\bfseries,
        rotate=90, anchor=center, inner sep=0pt},
    glyph/.style={fill=#1, text=white, font=\bfseries\scriptsize,
        inner sep=0pt},
    flow/.style={->, black!60, line width=0.5pt},
    dots/.style={font=\normalsize, text=black!55},
]
\def\xa{1.75}\def\xb{3.55}\def\xd{4.95}\def\xk{6.55}\def\xc{4.15}

\def\bcband#1#2#3#4{%
    \fill[#1!7] (0,#2) rectangle (7.7,#3);
    \fill[#1] (0,#2) rectangle (0.42,#3);
    \node[text=white, font=\sffamily\scriptsize\bfseries, rotate=90]
        at (0.21,{(#2+#3)/2}) {#4};}
\bcband{bcS1}{0.1}{-1.0}{1\;Fork}
\bcband{bcS2}{-1.1}{-3.3}{2\;Roll out}
\bcband{bcS3}{-3.4}{-5.05}{3\;Score}
\bcband{bcS4}{-5.15}{-6.75}{4\;Commit}

\node[card=bcS1, minimum width=3.4cm, minimum height=0.64cm] (state) at (\xc,-0.5)
    {\textbf{Decision state $s$}\\[-0.5pt]
     fork; roll out frozen $\pi$ for $H$ steps};

\foreach \x/\k/\n in {\xa/1/1, \xb/2/2, \xk/K/k}{
    \node[glyph=bcS2, regular polygon, regular polygon sides=6,
          minimum size=4.6mm] (p\n) at (\x,-1.62) {$\pi$};
    \node[card=bcS2] (b\n) at (\x,-2.65)
        {\textbf{Branch $\tau_{\k}$}\\[-0.5pt]
         first action {\color{bcAct}$a_{\k}$}\\[-0.5pt]
         randomness {\color{bcReal}$\xi_{\k}$}};
    \draw[flow] (state.south) -- (p\n.north);
    \draw[flow] (p\n) -- (b\n);
}
\node[dots] at (\xd,-2.65) {$\cdots$};

\foreach \x/\k/\n in {\xa/1/1, \xb/2/2, \xk/K/k}{
    \node[glyph=bcS3, rounded corners=2.2pt, minimum width=4.2mm,
          minimum height=3.2mm] (v\n) at (\x,-3.78) {$v$};
    \node[card=bcS3, minimum height=0.62cm] (s\n) at (\x,-4.55)
        {{\tiny full-branch score}\\[-0.5pt] $O_v(\tau_{\k})$};
    \draw[flow] (b\n) -- (v\n);
    \draw[flow] (v\n) -- (s\n);
}
\node[dots] at (\xd,-4.55) {$\cdots$};

\node[glyph=bcS4, circle, minimum size=5mm, font=\bfseries\tiny] (mx)
    at (\xc,-5.5) {max};
\node[card=bcS4, minimum width=3.4cm, minimum height=0.62cm, line width=0.8pt]
    (commit) at (\xc,-6.3)
    {\textbf{Selected branch} $\tau^\star=\arg\max_k O_v(\tau_k)$\\[-0.5pt]
     resume from its endpoint};
\foreach \n in {1,2,k}{\draw[flow] (s\n.south) -- (mx);}
\draw[flow] (mx) -- (commit);

\begin{scope}[shift={(0,-6.9)}]
    \fill[bcAct!8] (0,0) rectangle (3.8,-0.92);
    \fill[bcAct] (0,0) rectangle (0.07,-0.92);
    \node[anchor=north west, align=left, inner sep=0pt] at (0.2,-0.1)
        {\textbf{\color{bcAct}Action selection}\\[1pt]
         {\color{bcAct}$a_i\neq a_j$}: branches differ\\[-0.5pt]
         in their first action};
    \fill[bcReal!8] (3.9,0) rectangle (7.7,-0.92);
    \fill[bcReal] (3.9,0) rectangle (3.97,-0.92);
    \node[anchor=north west, align=left, inner sep=0pt] at (4.1,-0.1)
        {\textbf{\color{bcReal}Realization selection}\\[1pt]
         $a_i=a_j$, {\color{bcReal}$\xi_i\neq\xi_j$}: same action,\\[-0.5pt]
         different stochastic outcome};
\end{scope}
\end{tikzpicture}}

    \caption{\textbf{Branch-and-commit with a frozen policy.}
    At a selected state, the operator forks the simulator, samples $K$
    candidate continuations from the unchanged policy, scores the complete
    continuations, and resumes execution from the endpoint of the selected
    branch. Gains can arise from \emph{action selection}, when branches
    begin with different actions, or from \emph{realization selection},
    when they begin with the same action but encounter different stochastic
    outcomes.}

    \Description{Flowchart. From the current simulator state, the simulator is forked and K continuations are sampled with the frozen policy. Each continuation is scored, and the agent commits to the highest-scoring one. Two notes below explain that gains come from action selection when branches start with different actions, and from realization selection when they start with the same action but see different stochastic outcomes.}
    \label{fig:branch_and_commit_overview}
\end{figure}

\paragraph{Branch-and-commit.}
At a decision state $s$, the operator generates $K \geq 2$ candidate
continuations. Each candidate samples an initial action from
$\pi(\cdot \mid s)$ and then follows $\pi$ for $H$ steps, producing
trajectories $\{\tau_1,\ldots,\tau_K\}$. The return observed during a
continuation is
\begin{equation}
    R(\tau_k)=\sum_{t=0}^{H-1}\gamma^t r_t^{(k)}.
\end{equation}
A verifier assigns a score to each continuation and selects
\begin{equation}
    \tau^*=\argmax_k O_v(\tau_k),
\end{equation}
where $O_v$ denotes the score assigned by verifier $v$. The operator then
commits the complete selected continuation and resumes execution from its
final state. Ideally, the verifier will choose the trajectory that maximizes both the obtained return and the maximum expected cumulative future discounted rewards, i.e., the value function of the state at the end of the trajectory. In practice, the exact value function is never known, and a learned approximation is not always available either.

Committing the continuation distinguishes this procedure from classical
rollout methods~\citep{tesauro1996online,bertsekas1997rollout}. Classical
rollout uses simulated continuations to select an action, which is then
executed in the environment. Branch-and-commit instead retains the realized
transitions and rewards of the selected continuation. This requires a
forkable and resumable environment.

\paragraph{Candidate dispersion.}
Let
\begin{equation}
    \bar R(s)=\frac{1}{K}\sum_{k=1}^{K}R(\tau_k)
\end{equation}
denote the mean return of the candidates generated at state $s$. We define
the \emph{candidate dispersion} at $s$ as
\begin{equation}
\label{eq:dispersion}
    \Delta(s)=\max_k R(\tau_k)-\bar R(s).
\end{equation}
When candidates are independent continuations of the frozen policy,
$\bar R(s)$ is the expected return from selecting a candidate uniformly.
Dispersion therefore measures how much better the best available
continuation is than a random candidate.
For example, if $K{=}3$ candidates obtain returns $2$, $5$, and $3$, then
$\bar R(s)=3.33$ and $\Delta(s)=5-3.33=1.67$: selecting the second
candidate gains $1.67$ over a random choice
(Appendix~\ref{app:worked_example} follows this decision to the
cell-level quantities used in our analysis).

The within-window gain obtained by verifier $v$ is
\begin{equation}
\label{eq:window_gain}
    g_v(s)=R(\tau^*)-\bar R(s).
\end{equation}
For any verifier,
\begin{equation}
\label{eq:dispersion_bound}
    g_v(s)\leq\Delta(s).
\end{equation}
Thus, dispersion describes the opportunity available in the candidate set,
while the verifier determines how much of that opportunity is captured. If
$\Delta(s)=0$, no verifier can improve the return within the branch window, but that does not mean that the trajectory is optimal, since the final state might have low expected future return or value.
Additional bounds relating verifier accuracy, branching width, and dispersion are provided in Appendix~\ref{app:proof_scoring_bound}.

\paragraph{Action and realization selection.}
The candidates can differ for two reasons. First, they may sample different
actions from $\pi(\cdot\mid s)$. In particular, selecting among these candidates can
therefore lead to a different initial action, which we call
\emph{action selection}. Second, candidates that sample the same action can
still differ because of stochastic environment transitions and subsequent
policy sampling. Selecting among these candidates is
\emph{realization selection}: the operator retains a favorable realization
of a decision the policy had already made. We distinguish these mechanisms
empirically by grouping candidates according to their first action and by
repeating the branching procedure with shared environment randomness
(Sections~\ref{subsec:decomposition} and~\ref{subsec:realization_control}).

\paragraph{Episode-level lift.}
Dispersion and within-window gain are measured during the $H$-step
continuations, but committing a branch can also affect later rewards. We
therefore evaluate the complete episode. Let $G$ denote episode return with
branching, $G_0$ the return of a matched no-branching run, and $N$ the number
of branching decisions in the episode. We define per-decision episode lift
as
\begin{equation}
\label{eq:lift}
    \ell=\frac{G-G_0}{N}.
\end{equation}

For each experimental condition $c$, we average dispersion over its branched
states to obtain $\Delta_c$ and average Equation~\ref{eq:lift} over its
episodes to obtain $\ell_c$. Our main empirical question is whether and how much local
candidate dispersion can predict the eventual episode-level benefit:
\begin{equation}
\label{eq:law}
    \ell_c \approx \beta_{v,e}\Delta_c,
\end{equation}
where the slope may depend on the verifier $v$ and environment $e$. Unlike
the within-window bound in Equation~\ref{eq:dispersion_bound}, this
relationship is not guaranteed: it depends on what happens after the
selected continuation ends. We test it directly in
Section~\ref{sec:experiments}.

\section{Experimental Methodology}
\label{sec:methodology}

\subsection{Environments and Policies}

Our primary environment is Craftax-Symbolic~\citep{matthews2024craftax}, an open-world survival environment with sparse rewards and a hierarchical crafting structure. We use the Medium difficulty setting for the main analysis and Easy and Hard as boundary conditions. We additionally evaluate on Asterix-MinAtar and two XLand-MiniGrid environments, R9-19x19 and DoorKey-8x8~\citep{nikulin2023xlandminigrid}. These environments differ in episode length, policy confidence, reward density, and environment stochasticity.

We study two frozen policy architectures: a PPO multilayer perceptron, denoted $\pi_{\mathrm{weak}}$~\citep{schulman2017proximal}, and a stronger GTrXL policy, denoted $\pi_{\mathrm{strong}}$~\citep{parisotto2020stabilizing}. The policies are not updated during any experiment. Craftax-Medium is the main experimental setting because it produces substantial variation in dispersion across policies and branching configurations. The remaining environments test whether the proposed account correctly identifies settings in which branching has little or no benefit.

\subsection{Branching Protocol}

At each branching decision, we independently sample $K$ continuations from the frozen policy. Each continuation samples an initial action from $\pi(\cdot\mid s)$ and follows the same policy for $H$ steps using an independent rollout seed. Unless otherwise noted, we use $K{=}7$ and $H{=}20$ in Craftax, and $H{=}15$ in the other environments. Additional conditions vary the branching width and rollout horizon.

Branching is invoked periodically rather than at every step. We select the triggering period separately for each environment to account for differences in episode length and to obtain comparable per-episode branching budgets. Section~\ref{sec:dispersion_gating} replaces periodic triggering with a learned gating rule while leaving candidate generation and evaluation unchanged.

For each branched decision, we record the initial observation, the first action and realized return of every continuation, the verifier score, and the selected candidate. Rewards occurring after termination within a continuation are masked. Full configurations, seeds, and implementation details are provided in Appendix~\ref{app:setup}.\ifdefined\submissionversion\footnote{Appendices~A--F are provided as supplementary material.}\fi

\subsection{Candidate Evaluation}

We consider two access regimes that differ in whether the frozen policy is accompanied by its learned value function.

\paragraph{Value-head access.}
When a learned value head $\hat V$ is available, each continuation is scored by combining its realized $H$-step return with the estimated value of its final state:
\begin{equation}
\label{eq:oracle_vf}
    O_{\mathrm{VF}}(\tau_k)
    =
    R(\tau_k)
    +
    \gamma^H \hat V\!\left(s_H^{(k)}\right).
\end{equation}
The operator selects
$\tau^*=\argmax_k O_{\mathrm{VF}}(\tau_k)$.
This verifier uses a component already available in the trained agent and requires no additional model training.

\paragraph{Policy-only access.}
When the value head is unavailable, we use a language model to compare textual descriptions of the candidate continuations. A renderer summarizes the states, actions, events, and rewards observed in each trajectory. The $K$ descriptions are presented together, and the language model returns the index of the preferred continuation:
\begin{equation}
\label{eq:oracle_llm}
    \tau^*_{\mathrm{LLM}}
    =
    \mathrm{LLM}\!\left(
        \mathrm{render}(\tau_1),\ldots,
        \mathrm{render}(\tau_K)
    \right).
\end{equation}
The language model does not receive the value head, policy logits, or underlying symbolic state vector. We evaluate GPT-4o, GPT-4o-mini, Qwen-2.5-14B-AWQ, and Qwen-2.5-72B-AWQ without task-specific training or in-context examples. Because this regime requires a meaningful textual renderer, we use it only in Craftax and Asterix. Renderer details and prompts are provided in Appendix~\ref{app:render_example}.

We also consider a model-free reference that selects the continuation with the highest realized $H$-step return. This rule uses no estimate of value beyond the branch window and provides a direct comparison with the learned verifiers.

\subsection{Experimental Conditions and Measures}

An experimental condition, or \emph{cell}, specifies the environment, policy, verifier, branching width, rollout horizon, and triggering schedule. The main analysis includes 103 cells, each aggregating at least three seeds and between 10 and 50 episodes per seed. Unless stated otherwise, candidates are independent stochastic continuations of the frozen policy. The engineered candidate-generation mechanisms examined in Section~\ref{subsec:b2} are analyzed separately.

For each cell $c$, we average Equation~\ref{eq:dispersion} over its branching decisions to obtain mean dispersion,
\begin{equation}
    \Delta_c
    =
    \operatorname*{avg}_{s\in c}\Delta(s).
\end{equation}
We compare every branching run with a matched no-branching run using the same initial conditions and seeds. Episode-level improvement is converted to per-decision lift using Equation~\ref{eq:lift}, and then averaged within the cell to obtain $\ell_c$.

Our primary analysis tests whether $\Delta_c$ predicts $\ell_c$. Craftax-Medium provides the main test because its conditions span a broad range of dispersion values. We estimate the relationship using a mixed-effects model with verifier as a random intercept. As a complementary descriptive analysis across all environments, we fit one through-origin slope per verifier,
\begin{equation}
\label{eq:verifier_slope}
    \hat\beta_v
    =
    \frac{\sum_{c\in C_v}\Delta_c\ell_c}
         {\sum_{c\in C_v}\Delta_c^2}.
\end{equation}
Because cells share environments, policies, and seeds, we do not treat the 103 cells as independent environment-level replications. We report within-environment analyses and alternative clustered specifications in Appendix~\ref{app:craftax_internal}.

Pairwise verifier comparisons use Welch's $t$-tests with Holm--Bonferroni correction. Confidence intervals for slopes and slope differences are obtained using 10,000 bootstrap iterations. When the claim concerns similarity between verifiers, we report the estimated difference and its confidence interval rather than interpreting a non-significant test as evidence of equivalence. Unless otherwise noted, tables report seed-pooled means with seed-level standard errors. Complete evaluation-pool provenance and robustness analyses are provided in Appendices~\ref{app:pools} and~\ref{app:anchor_robustness}.

\section{When Does Branching Help?}
\label{sec:experiments}

Branching can improve return only if the candidate set contains meaningfully different outcomes. We test whether candidate dispersion, measured during the $H$-step branch window, predicts the episode-level lift obtained by committing a selected continuation.

\subsection{Dispersion Predicts Episode-Level Lift}
\label{subsec:lift_dispersion}

Craftax-Medium provides the main test of this relationship. Its 44 experimental conditions cover both policies, all five verifier classes, and a range of branching configurations, producing mean dispersion values from $0.09$ to $0.99$. Across these conditions, per-decision episode lift increases approximately linearly with dispersion. The mixed-effects model described in Section~\ref{sec:methodology} estimates a slope of $1.07 \pm 0.15$, with a 95\% confidence interval of $[0.78,1.36]$. Alternative specifications produce slopes between $0.90$ and $1.20$ (Appendices~\ref{app:craftax_internal} and~\ref{app:anchor_robustness}).

Part of this relationship follows directly from the definition of dispersion. Within the branch window, the selected continuation cannot outperform the best candidate, and its gain is therefore bounded by $\Delta(s)$ (Equation~\ref{eq:dispersion_bound}). However, the relationship with complete episode return is not guaranteed. After the selected continuation ends, the frozen policy continues acting from the state that the continuation reached. The effect of this state on later rewards may strengthen, eliminate, or reverse the immediate gain. The estimated slope is consistent with this effect preserving or strengthening the gain on average. Its point estimate exceeds one, which would indicate that the selected continuations also end in states from which the frozen policy earns more, although the confidence interval includes one (Appendix~\ref{app:slope_interpretation}).

The variation in dispersion also does not arise only because some conditions use more candidates. Increasing $K$ mechanically increases the maximum return of the candidate set and therefore its dispersion. We consequently repeat the analysis at the fixed operating budget $K{=}7$, $H{=}20$. The estimated slope is $1.20$, using the 42\% of dispersion variation that remains when the branching budget is held constant. The relationship is therefore not explained solely by varying the number or length of the continuations. Additional analyses across branching widths and rollout horizons are reported in Appendices~\ref{app:k_saturation} and~\ref{app:kh_invariance}.

Figure~\ref{fig:scaling_law_fit} summarizes the results across all 103 experimental conditions. For each verifier, the figure compares the lift predicted from dispersion with the observed episode-level lift. The pooled through-origin fits give $R^2_{\mathrm{origin}}=0.85$. We treat this value as descriptive rather than as evidence from 103 independent replications: cells share environments, policies, and seeds, and 41 cells are concentrated near the origin. Removing these near-zero cells changes the estimated slopes by at most $0.02$, although it changes the $R^2$ substantially (Appendix~\ref{app:anchor_robustness}). The main evidence for the relationship therefore comes from the slope estimated within Craftax-Medium, where dispersion varies over a wide range.

\begin{figure}[t]
    \centering
    \includegraphics[width=\columnwidth]{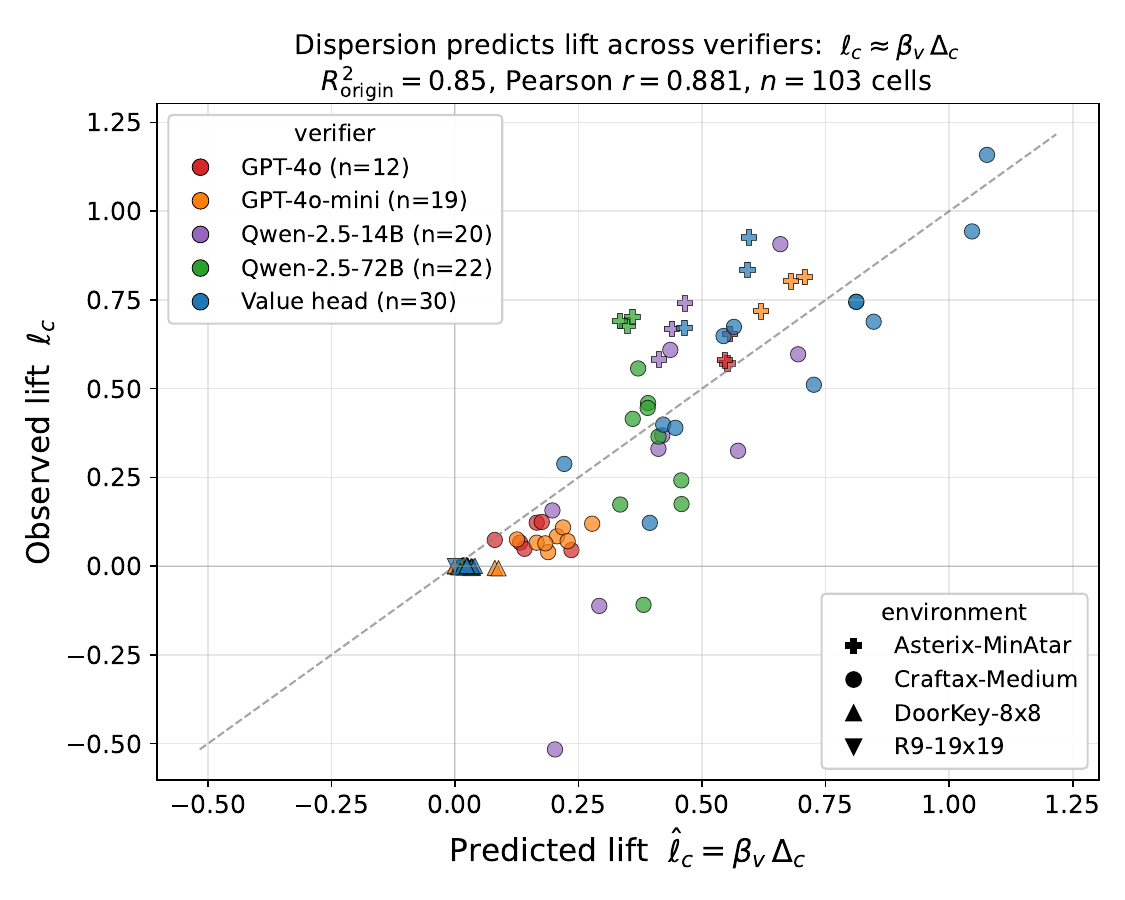}
    \caption{\textbf{Candidate dispersion predicts episode-level lift.}
    Predicted lift $\hat{\ell}_c=\hat{\beta}_v\Delta_c$ versus observed lift
    $\ell_c$ across 103 conditions. Colors indicate verifier classes and
    markers indicate environments. The dashed line shows equality between
    predicted and observed lift. Slopes are fitted separately for each
    verifier through the origin.}
    \Description{Scatter plot of predicted lift against observed lift for 103 conditions, colored by verifier class and shaped by environment. Points lie close to a dashed equality line, with a separate fitted line through the origin for each verifier.}
    \label{fig:scaling_law_fit}
\end{figure}

\subsection{Low-Dispersion Settings Provide Little Benefit}
\label{subsec:dispersion}

Before measuring the branching gains, we identified R9-19x19 and Craftax-Easy as settings in which the candidate continuations should have little dispersion. Both behave as predicted. R9-19x19 has effectively zero dispersion and branching produces no improvement. Craftax-Easy also has low dispersion and shows only a $2.1\%$ increase in return, within seed-level variation around zero. Increasing its branching frequency by a factor of three does not produce a benefit, showing that additional computation cannot compensate for a candidate set with little useful variation.

DoorKey-8x8 provides a related boundary case. The frozen policy already achieves an average reward of $0.97$, leaving little room for the sampled continuations to differ in useful ways. Its conditions also cluster near zero dispersion and zero lift. These results arise for different reasons (limited candidate variation in R9-19x19 and Craftax-Easy, and near-saturated policy performance in DoorKey) but lead to the same conclusion: when the candidates do not contain a substantially better continuation, branching provides little benefit.

The comparison across Craftax difficulty levels provides the clearest within-environment illustration. Under $\pi_{\mathrm{strong}}$, Craftax-Easy has mean dispersion $\Delta_c=0.04$ and produces essentially no lift. Dispersion increases substantially on Craftax-Medium, where branching improves return by $12.8\%$ at the main operating point. On Craftax-Hard, dispersion falls again to $0.064$ and branching reduces return by $3.9\%$. At such low dispersion, the small potential selection gain does not compensate for committing a sampled continuation that may move the agent to a less favorable state.

Asterix provides a second positive-dispersion environment, but its conditions cover a relatively narrow dispersion range. Branching improves return there, but the data support a positive operating level rather than an independent estimate of the dispersion--lift slope (Appendix~\ref{app:craftax_internal}). Overall, the cross-environment results are consistent with dispersion as a useful screening measure: it identifies settings where branching can help and, equally importantly, settings where additional test-time computation is unlikely to be useful.

\section{What Produces the Gain?}
\label{sec:mechanism}

Candidate continuations can differ because the policy samples different
actions or because stochastic transitions produce different outcomes after
the same action. We next examine which source of variation accounts for the
dispersion observed in our experiments.

\subsection{Most Branches Begin with the Same Action}
\label{subsec:decomposition}

We first group candidate continuations according to their initial action.
Variation between the mean returns of these groups reflects differences
associated with the policy's first action, while variation within a group
reflects what happens after that action is fixed.

Under $\pi_{\mathrm{strong}}$ on Craftax-Medium, all $K{=}7$ candidates
sample the same first action at 84.9\% of branching decisions. Moreover,
only 3.1\% of the observed branch-return variance lies between first-action
groups. This share remains in the single digits across all branching widths
we examine. Thus, in most decisions, branch-and-commit does not choose
between different initial actions. It chooses among different continuations
of the same action.

The result depends on the policy. Under $\pi_{\mathrm{weak}}$ on
Craftax-Medium, all candidates select the same first action at only 15.1\%
of decisions, and 40.5\% of the variance lies between action groups.
Asterix, which also uses the weaker policy, shows a similar between-action
share of 40.0\%. Branching therefore has a larger action-selection component
when the policy is less concentrated. The complete decomposition is
reported in Table~\ref{tab:action_vs_noise}.

\begin{table}[t]
    \centering
    \caption{\textbf{Evidence for realization selection on
    Craftax-Medium.} ``Independent'' uses independent environment
    randomness across branches; ``shared RNG'' uses common environment
    randomness while keeping policy sampling independent.}
    \label{tab:mechanism_summary}
    \small
    \setlength{\tabcolsep}{4pt}
    \begin{tabular}{@{}lcc@{}}
        \toprule
        Measure & Independent & Shared RNG \\
        \midrule
        Same initial action & 84.9\% & --- \\
        Between-action variance & 3.1\% & --- \\
        Mean dispersion $\Delta$ & 0.455 & 0.142 \\
        Zero-dispersion decisions & 58.5\% & 84.8\% \\
        \bottomrule
    \end{tabular}
\end{table}

\subsection{Controlling Environment Randomness}
\label{subsec:realization_control}

The preceding analysis is observational: grouping candidates by first
action does not directly intervene on the source of their later differences.
We therefore repeat the main Craftax-Medium condition using common
environment randomness across branches. At every rollout step, all branches
receive the same environment random draws, while action sampling remains
independent. Candidates can still diverge when the policy selects different
actions, but not because identical action sequences receive different
environment outcomes.

Sharing environment randomness reduces mean dispersion from $0.455$ to
$0.142$, a difference of $-0.313$ with a 95\% confidence interval of
$[-0.453,-0.173]$. It also increases the proportion of decisions with no
dispersion from 58.5\% to 84.8\% (Table~\ref{tab:mechanism_summary}). The reduction occurs in every seed.
Approximately 69\% of the dispersion observed with independent branches
therefore disappears when the environment's stochastic outcomes are shared.

The residual dispersion is larger than the 3.1\% of variance attributed to
the first action because the two analyses measure different quantities. The
variance decomposition isolates only the first action, whereas policy
sampling remains independent throughout the full $H$-step continuation in
the common-random-numbers condition. Candidates may therefore diverge
through actions selected later in the branch. Both analyses nevertheless
lead to the same conclusion: under the stronger policy, most of the
available variation comes from alternative realizations rather than from
the initial action selected by the policy. Full per-seed results are
reported in Appendix~\ref{app:crn_control}.

Episode return moves in the direction this account predicts. With
independent randomness, branching returns $36.00$; with shared randomness,
it returns $31.53$, against a no-branching baseline of $32.09$. Removing
the realizations from the candidate set therefore removes the lift. With
30 episodes per condition we read this as a direction rather than an
effect estimate, since the two conditions visit different later states once
the selected continuation changes. The intervention's primary role is to
identify the source of candidate dispersion.

\subsection{Comparison with Classical Rollout}

The distinction between action and realization selection also explains why
branch-and-commit differs from classical rollout. A rollout algorithm groups
simulations by their initial action, estimates the value of each action from
its rollouts, and then executes the selected action. If every candidate
samples the same first action, the rollout algorithm has no choice to make.
Branch-and-commit can still benefit because it retains the complete
realization of the selected continuation.

We quantify this difference using the same logged continuations. For each
decision, we compare the gain from selecting the candidate with the highest
return with the gain from selecting the action whose candidates have the
highest mean return. At the main Craftax-Medium operating point, classical
rollout captures only 10.6\% of the within-window gain available to
branch-and-commit, with a 95\% bootstrap interval of $[6.4\%,16.5\%]$. The
fraction is substantially larger for the weaker policy and in Asterix,
where action choice accounts for more of the candidate variation
(Appendix~\ref{app:rollout_contrast}).

This comparison uses identical logged candidates and isolates the mechanism;
it is not a matched-budget comparison between two deployed algorithms. A
rollout algorithm would execute the selected action and re-plan from the
resulting state, whereas branch-and-commit resumes from the end of the
selected continuation. Comparing their full episode returns would therefore
require additional experiments. The present result establishes the narrower
point needed here: in our main setting, most of what branch-and-commit gains
would be discarded by a method that retains only the selected action.

The same view covers restarting whole episodes. Uniform branching adds
$7\times7\times20=980$ simulator steps per episode, and dispersion gating
reduces this to about $470$. A comparable budget could instead simulate
several complete episodes from the initial state and keep the best. This is
branch-and-commit with a single decision at the start of the episode and a
horizon equal to its length, the limiting case of this design space and the
decision with the highest available dispersion. Our account therefore
predicts a substantial gain from it, and the caveat of
Section~\ref{sec:discussion} applies to it directly: it raises
reported return without changing the policy. It also requires simulating a
complete episode before acting, which rules it out for continuing tasks or
when actions must be taken online. Comparing these allocations at a matched
simulation budget is future work.

Overall, the improvement from branch-and-commit should not be interpreted
primarily as better action selection. Under the stronger policy, branching
mostly selects a favorable outcome of a decision the policy had already
made. This explains both the value of fork-and-resume access and the limited
scope of the method outside digitally instantiated environments.

\section{How Should the Computation Be Spent?}
\label{sec:allocation}

The previous sections show that branching gains depend on the variation
available in the candidate set and that, under the stronger policy, most of
this variation comes from environment realizations. We next consider three
choices available to the operator: which verifier to use, how to generate
the candidates, and which decisions to branch on.

\subsection{Competent Verifiers Achieve Similar Episode Returns}
\label{sec:investigation}

We first compare the value-head verifier with the four language-model
verifiers and with a model-free rule that selects the candidate with the
highest realized $H$-step return. On Craftax-Medium, value-head selection
increases episode return from the no-branching baseline of $32.09$ to
$36.21$. GPT-4o reaches $35.84$, and the model-free realized-return rule
reaches $35.40$. As summarized in Table~\ref{tab:verifier_comparison}, three verifiers
with substantially different information and computational requirements
produce similar episode returns.

We state the comparison between the value head and GPT-4o as a bounded
difference rather than interpreting a non-significant test as evidence of
equivalence. The estimated difference is $0.37$, with a 90\% confidence
interval of $[-0.16,0.89]$. The largest difference compatible with this
interval is therefore $0.89$ return units, or 22\% of the gain obtained by
branching. GPT-4o-mini also falls within this range. The two Qwen verifiers
have lower point estimates: Qwen-14B and Qwen-72B are $0.67$ and $1.43$ return
units below the value head, respectively, and both have significantly lower
dispersion--lift slopes (Appendix~\ref{app:pairwise_beta}). Our data do not
resolve the ordering between the two Qwen models. Randomly selecting a candidate produces a
return of $32.00$, nearly identical to not branching, showing that candidate
generation alone provides no benefit without meaningful selection.

\begin{table}[t]
    \centering
    \caption{\textbf{Verifier comparison on Craftax-Medium.}
    Episode-return point estimates use the main $K{=}7$ setting.
    Uniform selection retains branching but selects a candidate
    at random.}
    \label{tab:verifier_comparison}
    \small
    \setlength{\tabcolsep}{6pt}
    \begin{tabular}{@{}lc@{}}
        \toprule
        Selection method & Episode return \\
        \midrule
        No branching          & $32.09$ \\
        Uniform selection     & $32.00$ \\
        Realized-return rule  & $35.40$ \\
        GPT-4o                & $35.84$ \\
        Value head            & $\mathbf{36.21}$ \\
        \bottomrule
    \end{tabular}
\end{table}

The similarity in episode return does not mean that the verifiers are
identical. Within Craftax-Medium, the value head captures a larger fraction
of the available dispersion than the language models. However, the
verifiers agree on most decisions, and their differences matter primarily
at the relatively rare decisions with high dispersion. Averaging over a
complete episode reduces the resulting return gap. Detailed comparisons,
including the verifier-specific slopes and disagreement analysis, are
reported in Appendices~\ref{app:within_arch_slope},
\ref{app:pairwise_beta}, and~\ref{app:llm_failure_modes}.

These results suggest that verifier quality matters up to a point: weak
verifiers can lose much of the available benefit, but replacing an already
competent verifier provides limited improvement at our operating budget.

\subsection{Increasing Dispersion Does Not Necessarily Improve Return}
\label{subsec:b2}

A second possibility is to modify candidate generation so that the branches
are more diverse. We evaluate five proposal mechanisms that encourage
different actions, including mechanically generated alternatives and
language-model-generated action prefixes. These mechanisms change mean
dispersion by as much as a factor of two, but none improves episode return
over the policy's own stochastic candidates. No individual comparison is
resolved with three seeds, but all five mechanisms fall below the stochastic
reference (one-sided sign test, $p=0.031$; Appendix~\ref{app:b2_proposer}).

This result does not contradict the relationship established in
Section~\ref{sec:experiments}. Dispersion measures the best candidate
relative to the mean of the candidate set. An off-policy proposer can raise
this difference by introducing poor candidates, or can select actions that
produce high short-term variation but leave the frozen policy in states
from which it performs poorly later. Dispersion is therefore useful for
characterizing candidates sampled from the deployed policy, but should not
be treated as an objective that can be increased without constraint.

\subsection{Predicting Where to Branch}
\label{sec:dispersion_gating}
\label{subsec:dispersion_targeted}

The remaining choice is where to spend the branching budget. Branching at a
low-dispersion state offers little benefit regardless of the verifier,
whereas branching at a high-dispersion state provides an opportunity for
selection. Because $\Delta(s)$ is observed only after generating the
candidates, it cannot be used directly as a gating rule. We instead predict
it from the current observation.

We collect states and their realized dispersion from runs using uniform
periodic branching and fit a linear ridge predictor
$\hat{\Delta}_{\theta}(s)$. Five-fold cross-validation on 1,780
Craftax-Medium decisions gives a Spearman correlation of $\rho=0.37$
between predicted and realized dispersion. At deployment, we branch only
when
\begin{equation}
    \hat{\Delta}_{\theta}(s)>\tau.
\end{equation}
The predictor is evaluated before any candidate is generated. Every skipped
decision therefore saves all $K$ rollouts and the verifier call. The policy,
candidate-generation procedure, and verifier remain unchanged.

The threshold was initially selected using the logged Craftax-Medium data,
so we evaluate it on twelve fresh seeds. On these held-out seeds,
dispersion-gated branching obtains a mean episode return of $35.17$ using
$3.34$ branching calls per episode. Uniform periodic branching obtains
$36.54$ using $7.0$ calls per episode. Relative to the no-branching
baseline, gating retains 69\% of the lift using 48\% of the calls
(Table~\ref{tab:gating_heldout}). The estimated improvement in lift per call
is $1.45\times$, although its 95\% bootstrap interval of $[0.90,2.28]$
includes parity. We therefore base the claim on retained lift and reduced
calls, rather than on a resolved efficiency ratio.

\begin{table}[t]
    \centering
    \caption{\textbf{Held-out evaluation of dispersion gating on
    Craftax-Medium.} Results use twelve fresh seeds. Percentages in
    parentheses are relative to uniform periodic branching. Uniform
    branching returns $36.54$ here and $36.21$ in
    Table~\ref{tab:verifier_comparison} because the two use different
    seed pools (Appendix~\ref{app:pools}).}
    \label{tab:gating_heldout}
    \footnotesize
    \setlength{\tabcolsep}{4pt}
    \begin{tabular}{@{}lccc@{}}
        \toprule
        Method & Return & Calls/ep. & Lift retained \\
        \midrule
        Uniform & $36.54 \pm 0.77$ & $7.00$ (100\%) & 100\% \\
        Dispersion-gated & $35.17 \pm 0.73$ & $3.34$ (48\%) & 69\% \\
        \bottomrule
    \end{tabular}
\end{table}

Offline analysis supports the same mechanism. The 30\% of decisions with
the highest predicted dispersion contain 68.0\% of the total observed
dispersion, compared with 30.4\% under random selection. Shuffling the
predicted scores within episodes preserves their distribution and the
number of triggered decisions but removes their alignment with the states;
the concentration then falls to the random baseline
(Appendix~\ref{app:gate_permutation}).

Gating is useful only when the environment contains a concentrated
high-dispersion subset. In Asterix, predicted dispersion is distributed
more evenly across decisions. The same procedure captures only
$1.14\times$ as much dispersion as random gating, compared with
$2.24\times$ in Craftax-Medium, and every tested gating configuration
underperforms uniform periodic branching at comparable or lower cost
(Table~\ref{tab:asterix_gating}). This negative result is consistent with
the proposed account: when useful dispersion is broadly distributed,
uniform branching is preferable to concentrating the budget on a small
set of states.

Together, the results suggest a simple allocation procedure. An operator
should first measure whether candidate dispersion is present. If it is
near zero, branching is unlikely to help. If dispersion is present but
distributed uniformly, the operator should use uniform triggering. When
dispersion is concentrated and predictable, gating can retain much of the
benefit while substantially reducing the number of branching calls.

\section{Related Work}
\label{sec:related}

\paragraph{Simulation-based planning.}
Branch-and-commit is closely related to rollout algorithms, sparse sampling,
Monte Carlo tree search, and model-predictive control, all of which use
simulated continuations to improve decisions at execution
time~\citep{tesauro1996online,bertsekas1997rollout,kearns2002sparse,
kocsis2006bandit,silver2010monte,chua2018deep}. Classical rollout methods
estimate the value of candidate actions from simulated trajectories and
then execute the selected action. Branch-and-commit differs in what is
retained: rather than using a continuation only to evaluate its first
action, it resumes execution from the final state of the selected
continuation. This difference makes environment realizations part of what
can be selected. Our analysis shows that, under a confident policy, this
realization-selection component accounts for most of the available gain.

Our setting also differs from model-based planning methods that learn or
approximate the environment dynamics~\citep{silver2016mastering,
schrittwieser2020mastering}. We assume direct access to a simulator that can
be forked and resumed, and do not learn a dynamics model. The assumption is
restrictive, but allows us to isolate how candidate variation, verifier
quality, and compute allocation affect the benefit of branching.

\paragraph{Test-time compute and candidate selection.}
Best-of-$N$ and search-based methods improve language-model outputs by
generating multiple candidates and selecting among
them~\citep{brown2024large,snell2025scaling,wang2022self,
yao2023tree,hao2023reasoning}. Because candidate quality is not directly
observable, this literature emphasizes learned reward models, process
verifiers, and other methods for identifying the best
candidate~\citep{stiennon2020learning,nakano2021webgpt,
cobbe2021training,lightman2023lets}. Prior work has accordingly studied how
verifier errors limit the benefit of additional
sampling~\citep{stroebl2024inference,huang2025bestofn}.

A simulator changes this balance because it directly reveals the rewards
received during each continuation. Verifier quality remains relevant for
estimating what happens beyond the simulated window, but it is not the only
constraint. A better continuation must also be present in the candidate
set. This distinction is related to the separation between candidate
coverage and verification in best-of-$N$ inference~\citep{brown2024large}.
Our dispersion measure captures the coverage available at an individual
decision state and places an upper bound on the gain obtainable within the
branch window.

Recent studies relate test-time scaling to variation in candidate quality
and develop adaptive methods that allocate computation based on predicted
benefit~\citep{setlur2025scaling,sareen2025putting,li2026slg,
li2026posttraining,qu2026adaptive,manvi2024adaptive,
li2025entropy}. The closest work gates language-agent computation using
disagreement among proposed actions~\citep{lee2026agentic}. Action
disagreement is insufficient in our setting: under the stronger policy,
all candidates choose the same first action at 84.9\% of decisions, even
though their returns can differ substantially. We instead predict
dispersion in candidate outcomes.

\paragraph{Metareasoning and simulation-budget allocation.}
Deciding whether additional computation is worthwhile is a central problem
in metareasoning and value-of-computation
research~\citep{russell1991principles,hay2012selecting}. Related work
allocates simulation budgets based on the variance of candidate outcomes
or the expected value of further sampling~\citep{chen2000ocba,
audibert2010best,bubeck2009pure}. State-importance measures have similarly
been used to decide when agents should request advice or which decisions
should be shown to people~\citep{torrey2013teaching,amir2018highlights}.
Our work applies this perspective to test-time branching for a frozen
policy. We first measure whether useful candidate variation exists and then
predict the states at which that variation is concentrated.

\paragraph{Language models as trajectory evaluators.}
Language models have been used to guide, train, or directly control
decision-making agents~\citep{ahn2022can,carta2023grounding,
du2023guiding,ma2023eureka,wang2023voyager}. In our work, the language model
does not act in the environment or generate the policy. It ranks textual
descriptions of trajectories produced by a separate frozen policy. This is
closest to work using foundation models as zero-shot evaluators of
actions~\citep{pignatelli2024assessing,rocamonde2024vision}, LLM-as-judge
methods~\citep{zheng2023judging}, and language-based evaluation of agent
trajectories~\citep{admoni2026from}. Our comparison with the agent's value
head and a realized-return rule tests when such a general-purpose evaluator
provides additional value once the simulator already exposes short-horizon
outcomes.

\section{Discussion and Conclusion}
\label{sec:discussion}

Our results characterize when branch-and-commit can improve the performance
of a frozen reinforcement learning policy and what that improvement
represents. Candidate dispersion measures the opportunity: when dispersion
is near zero, branching provides little or no benefit; when dispersion is
substantial, it predicts the episode-level improvement. Under the stronger
Craftax policy, the gain comes primarily from selecting a favorable
realization of an action the policy had already chosen. Branch-and-commit
improves the behavior of the combined policy--environment system, but it
does not improve the policy itself.

This distinction matters for evaluation. Applying branch-and-commit inside
an evaluation harness can increase the reported return of a checkpoint
without changing its action policy. Such a result should not be presented
as evidence that the underlying policy has improved, and evaluation should
report whether
gains come from selecting different actions or from retaining favorable
environment realizations. Branch-and-commit is most naturally viewed as a
deployment procedure for digitally instantiated environments in which
selecting a simulated state is an allowed operation, rather than as a
method for comparing the quality of frozen policies.

For the allocation of computation, a more complex verifier adds little once
the verifier is sufficiently accurate, and off-policy candidate generation
does not improve return because it can introduce poor candidates or move
the policy to unfavorable later states. In contrast, predicting where
naturally occurring dispersion is high reduces the number of branching
calls while retaining much of the benefit. The practical procedure suggested by our results
is therefore to first measure whether useful dispersion exists, and then
determine whether it is concentrated in states that can be identified
before branching. These conditions hold in digital deployments such as emulated
games and sandboxed software environments, where the operator can resume
from any simulated state.

\paragraph{Limitations.}
The most important limitation is the fork-and-resume assumption. The method
applies when the environment itself is digital and a selected simulated
continuation can become the subsequent environment state. It does not apply
directly to physical-world control, where simulations can inform action
selection but cannot determine which environmental realization occurs. An operator without resume access can
retain only the selected action, which under a confident policy removes
most of the available gain.

The empirical evidence is also concentrated in one environment.
Craftax-Medium provides the dispersion range needed to estimate the
dispersion--lift relationship; Asterix covers a narrow range,
and the remaining settings mainly serve as low-dispersion boundary
conditions, so the 103 cells are not independent replications across 103
tasks. Evaluating the account in additional resumable environments,
particularly environments with delayed rewards not visible within the
branch window, is an important direction for future work.

Our verifier comparisons and gating results also have limited statistical
resolution: the data do not reliably order similar language models, and
the confidence interval for gating's lift-per-call ratio includes parity,
so the result supports a trade-off between return and computation rather
than a precisely estimated efficiency multiplier.
Our gating experiments use the inexpensive value-head verifier; because the
gate fires before candidates are generated, savings grow with verifier
cost. We also do not compare against changing how the frozen policy samples
actions, for example by lowering its temperature. Such changes may recover
part of the action-selection gain, but they cannot select among
realizations, which account for most of the gain under the stronger policy.
Finally, selecting for high realized returns can also change the
distribution of outcomes: in an exploratory comparison across pools of
different composition, catastrophic episodes rise from 3.0\% to 6.1\%
(Appendix~\ref{app:tail_risk}), so applications with asymmetric costs
should evaluate more than the mean.

In summary, test-time branching is useful when the policy and environment
produce continuations worth selecting among. In the settings we study,
candidate dispersion provides a simple way to identify this opportunity,
to distinguish action selection from realization selection, and to decide
where additional simulation is most valuable. Future work could use
predicted dispersion to decide where to branch during training, and test
the account in other resumable domains such as sandboxed software agents.

\fi

\ifdefined\preprintversion\else
\begin{acks}
\end{acks}
\fi

\bibliographystyle{ACM-Reference-Format}
\bibliography{bibs}

\ifdefined\usefullaux\else
\makeatletter
\@for\@tmpc:={table,figure,equation,algorithm,theorem,lemma,proposition,corollary,definition,assumption,remark,question}\do{%
  \@ifundefined{c@\@tmpc}{}{%
    \immediate\write\@auxout{\string\newlabel{appstart@\@tmpc}{{\the\csname c@\@tmpc\endcsname}{}{}{}{}}}}}
\makeatother
\appendix

\fi

\end{document}